\documentclass[]{fairmeta}
\usepackage{wrapfig}    % Added for wrapfigure
\usepackage{latexsym}
\usepackage{hyperref}
\usepackage{url}
\usepackage[normalem]{ulem}
\useunder{\uline}{\ul}{}
\usepackage[export]{adjustbox}
\usepackage{graphicx}
\usepackage{rotating}
\usepackage{subcaption}

\usepackage{xcolor} % Required for defining colors
\usepackage{tcolorbox} % Required for creating colored boxes

\usepackage{graphicx}
\usepackage{amsmath}
\usepackage{amssymb}
\usepackage{booktabs} % For better table rules
\usepackage{multirow}
\usepackage{xcolor} % For text coloring
\usepackage{colortbl} % For table cell coloring (if needed)
\usepackage{caption} % For better caption control
\usepackage{subcaption} % For subfigures/subtables if needed
\usepackage{natbib} % For \[cite]p
\usepackage{hyperref} % For clickable references/links
\usepackage{float} % For [H] placement specifier
\usepackage{tabularx} % For flexible width tables
\usepackage{enumitem} % For customized lists
\usepackage{geometry} % Adjust margins if needed
\usepackage{tikz} % For potential diagrams if needed
\usepackage{pgfplots} % For plotting directly in LaTeX if needed
\usepackage{siunitx} % For consistent number/unit formatting
\usepackage{cleveref} % For smart cross-referencing (\cref)

\definecolor{forestgreen}{rgb}{0.13, 0.55, 0.13}

\usepackage{hyperref}

\usepackage[T1]{fontenc}
\usepackage[utf8]{inputenc}
\usepackage{microtype}
\usepackage{inconsolata}
\usepackage{graphicx}

\title{Demystifying Synthetic Data in LLM Pre-training: \\A Systematic  Study of Scaling Laws, Benefits, and Pitfalls}

\author[1,2,*]{Feiyang Kang}
\author[1]{Newsha Ardalani}
\author[1]{Michael Kuchnik}
\author[1]{Youssef Emad}
\author[3,*]{Mostafa Elhoushi}
\author[4,*]{Shubhabrata Sengupta}
\author[1]{Shang-Wen Li}
\author[1]{Ramya Raghavendra}
\author[2]{Ruoxi Jia}
\author[1]{Carole-Jean Wu}

\affiliation[1]{FAIR at Meta}
\affiliation[2]{Virginia Tech}
\affiliation[3]{Cerebras Systems}
\affiliation[4]{Independent consultant}

\contribution[*]{Work done at Meta}
\abstract{Training data plays a crucial role in Large Language Models (LLM) scaling, yet high quality data is of limited supply. Synthetic data techniques offer a potential path toward sidestepping these limitations.We conduct a large-scale empirical investigation (>1000 LLMs with >100k GPU hours) using a unified protocol and scaling laws, comparing natural web data, diverse synthetic types (rephrased text, generated textbooks), and mixtures of natural and synthetic data. Specifically, we found pre-training on rephrased synthetic data \textit{alone} is not faster than pre-training on natural web texts; while pre-training on 1/3 rephrased synthetic data mixed with 2/3 natural web texts can speed up 5-10x (to reach the same validation loss) at larger data budgets. Pre-training on textbook-style synthetic data \textit{alone} results in notably higher loss on many downstream domains especially at small data budgets. ``Good'' ratios of synthetic data in training data mixtures depend on the model size and data budget, empirically converging to $\sim$30\% for rephrased synthetic data. Larger generator models do not necessarily yield better pre-training data than $\sim$8B-param models. These results contribute mixed evidence on “model collapse" during large-scale single-round (n=1) model training on synthetic data–training on rephrased synthetic data shows no degradation in performance in foreseeable scales whereas training on mixtures of textbook-style pure-generated synthetic data shows patterns predicted by “model collapse". Our work demystifies synthetic data in pre-training, validates its conditional benefits, and offers practical guidance. }

\date{\today}
\correspondence{Feiyang Kang at \email{fyk@vt.edu}, Newsha Ardalani \email{new@meta.com}}

\begin{document}

\maketitle

\section{Introduction}
\label{sec:introduction}

The remarkable advancements in Large Language Models (LLMs) are closely tied to the scale and, critically, the quality of their training data. As computational demands for training state-of-the-art models escalate and the finite nature of high-quality natural text becomes increasingly apparent \citep{villalobos2024position}, significant interest has turned towards \textit{synthetic data} \citep{benallal2024cosmopedia,eldan2023tinystories,patel2024datadreamer,chen2024diversity,long2024llms,thrush2024improving,havrilla2024surveying,maini2024rephrasing,li2023textbooks,abdin2024phi,javaheripi2023phi,cheng2024instruction,gu2023pre}. Defined as text generated by pre-existing models or automated pipelines, synthetic data presents a compelling potential avenue for augmenting---or perhaps eventually replacing---traditional human-generated corpora during the foundational pre-training phase.

While the utility of synthetic data is increasingly established in post-training stages like instruction-tuning and alignment \citep{alpaca, li2023self, ge2024scaling}---where objectives are targeted and natural data may be scarce—its role and effects during the crucial pre-training phase remain largely uncharacterized and poorly understood \citep{liu2024best}. This knowledge gap represents a significant barrier to optimizing LLM development pipelines and motivates fundamental questions:

\begin{itemize} % [leftmargin=*, itemsep=1pt, topsep=2pt]
\item[\textbf{(RQ1)}] Can synthetic data effectively enhance LLM pre-training performance at large data scales and under what conditions?

\item[\textbf{(RQ2)}] How do different types and generation methodologies for synthetic data influence pre-training dynamics and scaling behavior?
\item[\textbf{(RQ3)}] What principles guide the effective deployment of synthetic data in pre-training, including ``good'' mixture ratios, the impact of generator model capabilities, and the statistics of the training corpus?
\end{itemize}

Despite the straightforward nature of these questions, clear answers remain elusive. This \textbf{ambiguity} stems from several factors. Firstly, the landscape is marked by \textit{inconsistent empirical findings and considerable methodological heterogeneity} \citep{long2024llms, liu2024best}. Proposed approaches often rely on bespoke setups, obscuring direct comparability and generalizability. %For example, Phi's attributes the success pretraining to the use of synthetic data \citep{li2023textbooks}---involving heavy manual effort in its curation, little information was revealed on the formatting, coverage, and distribution of the synthetic data. 
Even for simple open-sourced methods, \citet{yang2024synthetic} reports models trained on synthetic data from \citet{maini2024rephrasing} saturates early on in continued pre-training without much performance gain. Secondly, synthetic data generation involves \textit{complex tradeoffs between targeted quality enhancement and broad distributional diversity}. % While synthetic data is believed to improve training data quality at the cost of diversity \citep{havrilla2024surveying}, recent studies also argue that data diversity alone can be predictive of model performance \citep{chen2024diversity}.
Recent studies present a contradiction regarding synthetic data: while some argue it improves training data quality at the expense of diversity \citep{havrilla2024surveying}, others suggest that diversity itself is a key predictor of model performance \citep{chen2024diversity}.
% Synthetic data risks a lack of richness compared to natural text and its generation with LLMs potentially introduces biases or artifacts. 
Thirdly, \textit{theoretical concerns persist}, notably ``model collapse'' from recursive training \citep{dohmatob2024tale,dohmatob2024strong}, even if catastrophic failures are not yet widespread. This confluence of \textbf{potential benefits, inconsistent evidence, methodological variance, and theoretical risks} underscores a critical need for systematic investigation. % The trajectory of models purportedly leveraging synthetic data, such as Microsoft's Phi series \citep{li2023textbooks,javaheripi2023phi,abdin2024phi} (reporting initial benefits potentially tempered by later optimization with more natural data), further highlights this complex narrative.

To address this critical gap and provide empirically grounded answers , we undertake a systematic, large-scale investigation into the role and effective use of synthetic data in LLM pre-training. Our study involves training over 1000 LLM variants (up to 3B parameters) on datasets comprising up to 200B tokens, utilizing over 100,000 GPU hours, enabling evaluation on the effect of model size and data regimes in its scaling laws.

Our principal findings reveal that:
\begin{enumerate}
\item Strategically incorporating specific synthetic data types can \textbf{significantly accelerate pre-training convergence}. Compared to pre-training on natural web texts, training on 1/3 rephrased synthetic data mixed with 2/3 natural web texts can speed up 5-10x (to reach the same validation loss) at larger data budgets.

\item However, the impact is \textbf{highly dependent on the synthetic data's type and characteristics}: Pre-training on rephrased synthetic data alone is not faster than pre-training on natural web texts; whereas pre-training on textbook-style synthetic data alone results in notably higher validation loss. 

\item ``Good'' ratios of synthetic data in training data mixtures are nuanced, varying with data type, target model scale, and budget, converging to $\sim3$0\% for rephrased synthetic data. Counter-intuitively, \textbf{larger or more capable generator models do not necessarily yield superior synthetic data} than $\sim$8B-param models for pre-training downstream models. 

\item We interpret the results with a focus on low-level statistics. Some unigrams that are frequent in test datasets but rare or absent in training datasets result in higher evaluation loss, whereas no single training set offers complete coverage. CommonCrawl has wider unigram coverage and the lowest KL-divergence to test datasets; however, it did not yield  superior performance, suggesting \textbf{"good" training data mixtures depend on factors beyond simple similarity and pointing to more complex diversity-quality trade-offs.}
\end{enumerate}

\section{Related Work}
\label{sec:related_work}

Our research intersects with several key areas in LLM development, particularly concerning the generation and use of synthetic data for pre-training, data mixture strategies, the application of scaling laws, and concerns around model collapse.

\paragraph{Synthetic Data in LLM Pre-training}
The utility of synthetic data is well-recognized in targeted later stages of training, such as instruction tuning \citep{alpaca}, alignment \citep{li2023self, ge2024scaling}, and increasingly for enhancing reasoning capabilities \citep{muennighoff2025s1}. \citet{meta2025llama} detail a dedicated ``mid-training'' stage using synthetic reasoning data, occurring after initial pre-training and prior to subsequent post-training with reinforcement learning (RL). In contrast, synthetic data's role in foundational pre-training for general capabilities is less established and characterized by varied approaches.  The Phi series \citep{li2023textbooks, javaheripi2023phi} pioneered the use of ``textbook-style'' synthetic data for pre-training production-grade models. \citet{abdin2024phi}, discussing later Phi models (e.g., Phi-4), argue this approach particularly boosts reasoning with large training budgets where natural web text offers diminishing returns, while also acknowledging potential downsides like limited factual grounding and increased hallucination risks. 
Other foundational pre-training explorations include \citet{eldan2023tinystories}'s story generation for smaller models, rephrasing existing texts \citep{maini2024rephrasing}, and employing diverse prompts for generation \citep{chen2024diversity, patel2024datadreamer, gu2023pre}.
Despite these explorations (see survey by \citet{havrilla2024surveying}), the landscape is characterized by methodological heterogeneity and sometimes conflicting outcomes (e.g., \citet{long2024llms} and \citet{liu2024best} on diversity and quality; \citet{yang2024synthetic} on saturation with rephrased data from \citet{maini2024rephrasing}). Our study differentiates itself by systematically evaluating multiple distinct synthetic data generation paradigms (rephrased web text, generated ``textbooks'') and their mixtures with natural data under a unified pre-training protocol and rigorous scaling law analysis across substantial data and model scales, aiming to clarify these ambiguities.

\paragraph{Data Curation, Mixing Strategies, and Scaling Laws}
Meticulous data curation and strategic mixing of diverse natural data sources are established as critical for LLM pre-training \citep{touvron2023llama,raffel2020exploring,penedo2024fineweb,xie2023doremi,ye2024data,liu2024regmix}. However, the systematic integration and scaling behavior of \textit{synthetic} data with natural corpora remain comparatively underexplored, despite promising initial findings suggesting benefits from such mixtures \citep{maini2024rephrasing,javaheripi2023phi}. Seminal scaling laws describe predictable relationships between LLM performance and factors like model size, dataset size, and compute \citep{kaplan2020scaling,hoffmann2022training} and have recently been extended to model natural data mixing strategies \citep{kang2024autoscale}. The scaling dynamics of pre-training specifically with synthetic data have been described as ``mysterious'' \citep{liu2024best}. For instance, models trained on certain synthetic data types can exhibit early performance saturation \citep{yang2024synthetic}, highlighting the need for a clearer understanding. Our work distinctively addresses these gaps by employing scaling law analysis as a primary \textit{evaluative tool}. We systematically investigate optimal mixture ratios of different synthetic data types with a natural web text baseline, examining how these ratios and overall pre-training effectiveness interact with synthetic data characteristics and varying data budgets. This approach aims to demystify the role of synthetic data in scalable LLM pre-training and provide empirically grounded guidance for its effective integration.

\paragraph{Model Collapse and Generational Degradation}
The prospect of training models predominantly on model-generated data has spurred theoretical investigations into ``model collapse'' or ``generational degradation,'' where recursive training might lead to a decline in model quality due to reduced diversity or amplified biases \citep{shumailov2023curse, dohmatob2024tale, dohmatob2024strong}. While these risks are highlighted in theoretical analyses and simulations, large-scale empirical evidence from practical pre-training scenarios, especially those still incorporating significant natural data, remains limited. Our study contributes direct empirical insights on ``model collapse'' during large-scale single-round (n=1) training on synthetic data by pre-training models on substantial datasets with varying proportions and types of synthetic data.

\section{Synthetic Data Generation Methods} 
\label{sec:synthetic_data_generation}

We investigate two distinct paradigms for generating synthetic data: web rephrasing and textbook-style pure synthetic data. These paradigms represent different philosophies for augmenting or replacing natural text in pre-training. % This choice reflects ongoing exploration into what constitutes effective pre-training data and is broadly consistent with categorization schemes discussed in recent analyses of synthetic data's role (e.g., technical report for Phi-4 \citep{abdin2024phi}). 

\subsection{Web Rephrasing}
\label{ssec:web_rephrasing}

% \paragraph{Motivation and Core Idea:} 
% This approach aims to enhance the quality, information density, or stylistic properties of readily available web text. Standard web corpora (e.g., CommonCrawl) are vast but often contain noise, redundancy, or suboptimal writing.
Inspired by techniques such as WRAP \citep{maini2024rephrasing}, web rephrasing leverages a pre-trained Language Model (LM) to refine existing web documents into a potentially more valuable pre-training resource. 
% This method transforms existing web content rather than generating text entirely from scratch.
% However, defining optimal rewriting objectives is challenging, as improvements in one aspect (e.g., stylistic consistency) might trade off against the naturalness or diversity inherent in the original data. Furthermore, the precise characteristics of demonstrably ``good'' pre-training data remain an open research question, motivating empirical investigations like ours.
% \paragraph{Implementation Mechanism:} 
In our study, we implemented web rephrasing by sampling documents from the CommonCrawl dataset. 
A pre-trained generator LM was prompted to rewrite these documents. Drawing inspiration from variants explored in \citet{maini2024rephrasing}, we generated two distinct styles intended to probe different potential benefits for pre-training:
\begin{itemize}[leftmargin=*, itemsep=1pt, topsep=2pt]
    \item \textit{High-Quality (HQ) Rephrasing:} Prompts instructed the generator model to rewrite the source text into clear, coherent, well-structured English, mimicking the style often found in high-quality sources like Wikipedia. This targets improving general text quality for foundational pre-training, akin to an aggressive data filtering or quality enhancement step.
    % While distinct from methods generating novel knowledge-intensive text from scratch (such as the textbook approach), 
    HQ rephrasing aims to increase the effective density and quality of information already present within the web corpus, aligning with the broader goal of improving data efficiency for pre-training.

    \item \textit{Question-Answering (QA) Rephrasing:} Prompts instructed the generator model to restructure the source text's information into a conversational question-answering format. This explores incorporating instruction-following or dialogue-like structures directly into the pre-training phase, potentially accelerating the development of alignment capabilities.
    This QA rephrasing approach relates to the growing interest in `instruction pre-training' \citep{cheng2024instruction},
    where downstream objectives like instruction-following or dialogue capabilities are incorporated early via synthetic data formatting. % The aim is to embed these skills more fundamentally during the foundational training stage.
\end{itemize}
% This process yielded two distinct synthetic corpora derived from rephrasing CommonCrawl, allowing us to compare the pre-training impact of these different stylistic transformations.

\subsection{Synthetic Textbooks (TXBK)}
\label{ssec:synthetic_textbooks}

% \paragraph{Motivation and Core Idea:}
This paradigm is driven by the hypothesis that dense, high-quality, educational content might be more compute-efficient for instilling certain capabilities (e.g., reasoning, coding, factual recall) compared to diffuse web text. The goal is to generate entirely novel content that mimics the structure, style, and information density of textbooks or high-quality educational materials. % This approach emphasizes creating structured, knowledge-rich, and pedagogically valuable content—such as textbook explanations, definitions, worked examples with explanations, or practice problems with solutions relevant to target topics—designed to teach fundamental concepts and problem-solving skills efficiently. This method, inspired by data generation strategies reportedly employed for models like the Phi series \citep{li2023textbooks}, 
% focuses on creating original synthetic text rather than rephrasing existing documents. While aiming for such structured, knowledge-rich content, this paradigm inherently faces challenges. These include maintaining stylistic diversity to avoid monotonous formatting, mitigating the risk of factual inaccuracies or hallucinations introduced by the generator model, and ensuring broad yet relevant topic coverage.
% \paragraph{Implementation Mechanism:}
For our experiments, we generated novel ``textbook-style'' documents. The generation process was seeded using keywords \textit{randomly sampled} from CommonCrawl to provide diverse starting points for various topics. A pre-trained generator LM (e.g., Mistral-7B) was then prompted using structured instructions to produce text resembling textbook chapters or tutorials on the seeded topic. These prompts explicitly encouraged the generation of clear explanations, definitions, illustrative examples (including code snippets with explanations where relevant), and potentially associated exercises or reasoning steps. Throughout the generation process, an emphasis was placed on striving for factual accuracy, coherence, and a clear pedagogical structure.

% This method aimed to create a purely synthetic corpus distinct in characteristics from the rephrased corpora described earlier. This allows us to evaluate the pre-training value of this "textbook-quality'' data generation paradigm. Quality control principles applied during our process focused on ensuring the generated content was coherent and relevant to the intended educational purpose and seeded topic.
% This methodology aligns with other recent efforts focused on generating large-scale, high-quality synthetic datasets for pre-training, such as the publicly released Cosmopedia dataset \citep{benallal2024cosmopedia}, 
% which also leverages LLMs to generate educational-style content, albeit potentially with different seeding strategies, generation constraints, or quality control mechanisms.

% \section{Empirical Results}

\section{Empirical Results}
\label{sec:empirical_results}
\subsection{Experimental Setup}
\label{ssec:experiment_setup}

We conduct large-scale pre-training experiments comparing models trained on: (1) a natural web corpus baseline, (2) purely synthetic datasets generated using our distinct Web Rephrasing and Synthetic Textbook paradigms (see Section~\ref{sec:synthetic_data_generation} for generation methodologies), and (3) various mixtures of natural and synthetic data. Approximately 600 LLM variants, with sizes up to 3 billion parameters, were trained on datasets of up to 200 billion tokens. This effort consumed over 70,000 GPU hours on NVIDIA A100 80G hardware.

\subsubsection{Datasets}
\paragraph{Natural Data Baseline:} Our natural data consists of English text sourced from unfiltered CommonCrawl (CC) dumps, processed via the RedPajama-v2 pipeline \citep{weber2024redpajama}. % TODO: Replace placeholder

\paragraph{Synthetic Data:} All synthetic datasets were generated using a Mistral-Instruct-7b-v0.1 model \citep{jiang2023mistral}, 
with input documents for rephrasing or seeding sampled from our unfiltered CC baseline. Standard generation sampling parameters and light heuristic post-filtering were applied. Generation details, prompt templates, and sample generations are provided in Appendix~\ref{app:generation_details}.

% TODO: Add Appendix label
The following synthetic types were produced:
\begin{itemize}[leftmargin=*, itemsep=1pt, topsep=2pt]
    \item \textit{Web Rephrasing (\citet{maini2024rephrasing}-like):} Generated by rephrasing CC documents using prompts optimized from \citet{maini2024rephrasing} to produce longer texts in two styles: \textbf{High-Quality (HQ)} $\&$ \textbf{Question-Answering (QA)}.
    \item \textit{Synthetic Textbooks (\cite{li2023textbooks}-like):} Novel multi-chapter \textbf{``textbooks'' (TXBK)} generated from CC-derived outlines, employing varied prompts targeting different audiences to encourage diversity. Each chapter averaged $\sim$450 tokens and often included exercises.
\end{itemize}

\paragraph{Training Data Mixtures:} For each synthetic data type (HQ, QA, or TXBK), we prepared datasets by concatenating and shuffling source data under these conditions: 100\% Natural (Unfiltered CC baseline); 100\% Synthetic (consisting entirely of one synthetic type: HQ, QA, or Textbook); 67\% Synthetic / 33\% Natural; 33\% Synthetic / 67\% Natural. 
For each experimental point (defined by model size and data budget), models were trained on these different mixtures, enabling direct comparison. $5+$ model variants were trained per condition for robust scaling law analysis.

\subsubsection{Models, Training, and Evaluation}
We use a standard decoder-only Transformer architecture based on Llama 3 \citep{grattafiori2024llama}, % TODO: Replace placeholder
with model sizes ranging logarithmically from 100M to 3B parameters. All models were trained from scratch using the Meta Lingua library \citep{meta_lingua} % TODO: Replace placeholder
on PyTorch \citep{paszke2019pytorch}. Following the line of research on scaling laws for LLMs \citep{kaplan2020scaling}, we define the size of all models being trained as their count of \textit{non-embedding parameters}, which are \textit{learnable parameters} in the model except for those associated with the input and output token embeddings. A consistent training regime was applied for fair comparison, including a cosine learning rate schedule (10\% warmup), a context length of 4096 tokens, an effective batch size of 1M tokens, and the Llama 3 pre-trained TikToken tokenizer (128k vocabulary) \citep{grattafiori2024llama}. 
The primary performance measure is per-token average perplexity (cross-entropy loss) calculated on a held-out diverse set of 14 non-code/math English text domains from The Pile \citep{gao2020pile} 
and the Wikitext-103 dataset \citep{merity2016pointer}, evaluated at the final checkpoint. Complete details are provided in Appendix~\ref{ssec:models_training_appendix}.

\subsubsection{Data Scaling}\label{ssec:data}
For a fixed model size (1B parameters), data scaling is modeled as:
$
    \hat{\mathcal{L}}(D)=\frac{B}{D^\beta}+E,
$
where $\mathcal{L}$ is validation loss, $D$ is training data budget, and $B, \beta, E$ are fitted coefficients.

\begin{figure}[htbp]
    \label{fig:enter-label}
 \begin{minipage}{0.50\textwidth}
        \centering
        \includegraphics[width=\linewidth]{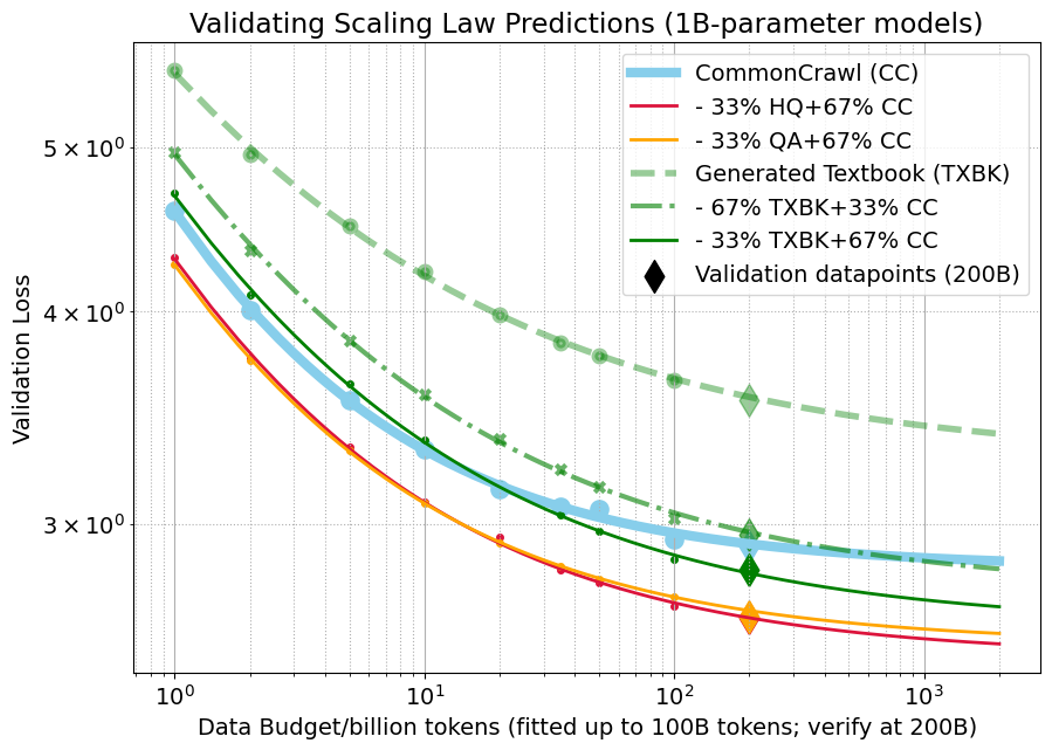}
        % \caption{Validation of the data scaling formula. Predictions for 200B tokens (fitted using up to 100B tokens) achieve an RMABE of 0.41\%. Solid dots display actual loss values while the fitted curves shows predicted loss. Validation datapoints are illustrated by diamond marks.}
        % \label{fig:dataval}
    \end{minipage}\hspace{-0em}
    \begin{minipage}{0.50\textwidth}
        \centering
        \includegraphics[width=\linewidth]{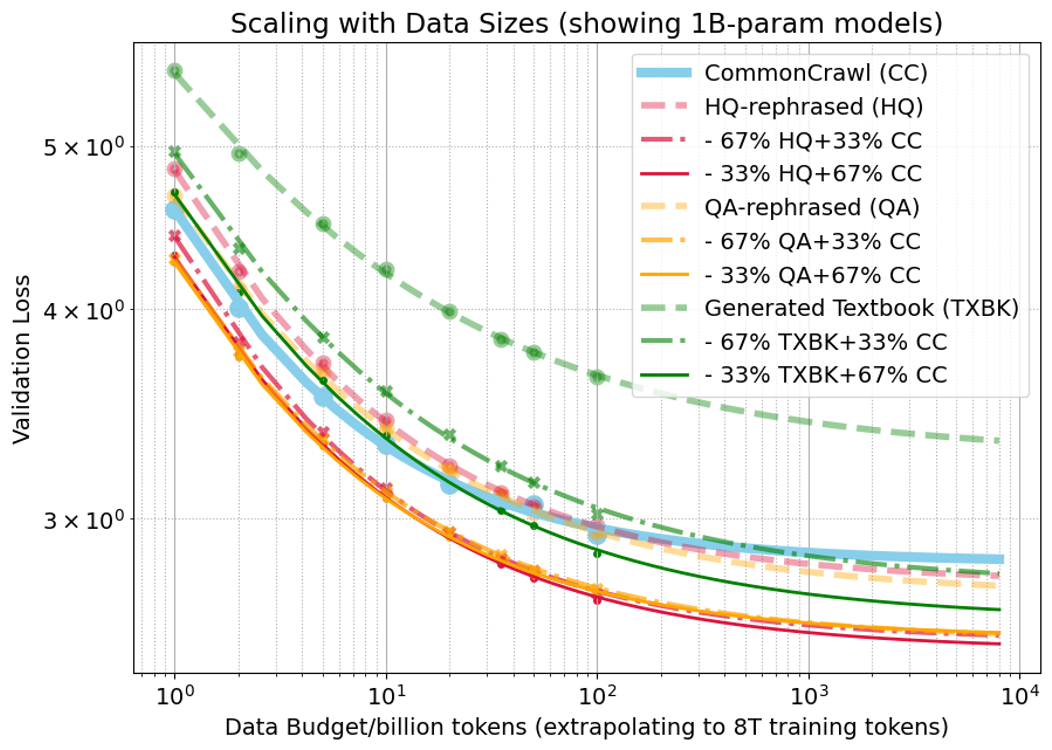}
        % \caption{Validation of the model scaling formula. Predictions for 3B-parameter models (fitted using up to 2B-parameter models) achieve an RMABE of 0.30\% on validation datapoints illustated with diamond marks.}
        % \label{fig:modelval}
    \end{minipage}
     \caption{Data Scaling. \textbf{Left:} Validation of the data scaling formula. Predictions for 200B tokens (fitted using up to 100B tokens) achieve an RMABE of 0.41\%. Solid dots display actual loss values while the fitted curves shows predicted loss. Validation datapoints are illustrated by diamond marks. \textbf{Right:} Extrapolated data scaling performance for 1B-parameter models across various data mixtures.}
     \label{fig:data-valscale}
\end{figure}

We trained 1B-parameter models on various data mixtures with data budgets from 1B up to 200B tokens. The scaling formula was fitted using data points up to 100B tokens; predictions were then validated on runs trained with 200B tokens. Six data mixtures (CommonCrawl, 33\% HQ + 67\% CC, 33\% QA + 67\% CC, Textbook (TXBK), 67\% TXBK + 33\% CC, and 33\% TXBK + 67\% CC) were trained to 200B tokens for this validation, as our HQ and QA synthetic datasets were limited to 100B tokens each.
The fit demonstrated high precision, as shown in Fig.~\ref{fig:data-valscale} (left), achieving a low Relative Mean Absolute Error (RMABE) of 0.41\% when predicting for 200B tokens.

% \begin{figure}[htbp]
%     \centering
%     \includegraphics[width=0.48\textwidth]{figs/modelval4.png}
%     \caption{Validation of the model scaling formula. Predictions for 3B-parameter models (fitted using up to 2B-parameter models) achieve an RMABE of 0.30\% on validation datapoints illustated with diamond marks.}
%     \label{fig:modelval}
% \end{figure}
% \begin{figure}[htbp]
%     \centering
%     \includegraphics[width=0.48\textwidth]{figs/dataval4.png}
%     \caption{Validation of the data scaling formula. Predictions for 200B tokens (fitted using up to 100B tokens) achieve an RMABE of 0.41\%. Solid dots display actual loss values while the fitted curves shows predicted loss. Validation datapoints are illustrated by diamond marks.} 
%     % {\color{red}\textit{(Addressing \texttt{\textbackslash mk} comment: The x-axis label in the figure could be clarified to explicitly state "Actual Loss at 200B Tokens" if it currently refers to the fitting range.)}}
%     \label{fig:dataval}
% \end{figure}

With \text{reasonably} reliable fits validated, we extrapolated data scaling (fitted up to 100B tokens) to predict loss for larger data budgets (up to 8T tokens), covering training regimes of state-of-the-art LLMs \citep{meta2025llama,deepseekai2024deepseekv3technicalreport}. % TODO: Replace placeholders
Key findings are presented in Fig.~\ref{fig:data-valscale} (right) and summarized below:
\begin{enumerate}[leftmargin=*, itemsep=1pt, topsep=2pt]
    \item \textbf{Pure synthetic data is not superior to CommonCrawl (CC):} Training solely on HQ or QA synthetic data does not \text{significantly} outperform training only on CC. Training only on TXBK performs notably worse than trainng on CC.
    \item \textbf{Mixtures outperform pure synthetic types:} Mixing any synthetic data type with CC substantially improves performance over using that synthetic type alone.
    \item \textbf{Rephrased data mixtures show low sensitivity to ratio (33\% vs. 67\% synthetic):} For HQ and QA, both 33\% and 67\% synthetic mixtures with CC yield similar performance.
    \item \textbf{Textbook mixtures favor less synthetic data:} For TXBK, a 33\% synthetic mixture significantly outperforms a 67\% mixture. The 33\% TXBK mixture surpasses pure CC performance after $\sim$20B tokens, while the 67\% TXBK mixture underperforms pure CC.
\end{enumerate}

% \begin{figure}[htbp]
%     \centering
%     \includegraphics[width=0.7\textwidth]{figs/datascale4.png}
%     \caption{Extrapolated data scaling performance for 1B-parameter models across various data mixtures.}
%     \label{fig:datascale}
% \end{figure}

\subsubsection{Model Scaling}\label{ssec:model}
For a fixed data budget (50B tokens), model scaling is modeled as:
$
    \hat{\mathcal{L}}(N)=\frac{A}{N^\alpha}+E, % Corrected D to N in the denominator
$
where $N$ is model parameter size, and $A, \alpha, E$ are fitted coefficients.

We trained models from 100M to 3B parameters for 50B tokens on all 10 data mixtures. The formula was fitted using models up to 2B parameters and validated on 3B-parameter models. This fit also proved highly precise (Fig.~\ref{fig:model-valscale}, left), with an RMABE of 0.30\% for 3B-parameter model predictions.
Extrapolating model scaling (fitted with models up to 3B parameters) to predict performance for larger models (up to 400B parameters) on a 50B token budget (Fig.~\ref{fig:model-valscale}, right) revealed several differences from data scaling patterns:

\begin{figure}[htbp]
    \label{fig:enter-label}
 \begin{minipage}{0.50\textwidth}
        \centering
        \includegraphics[width=\linewidth]{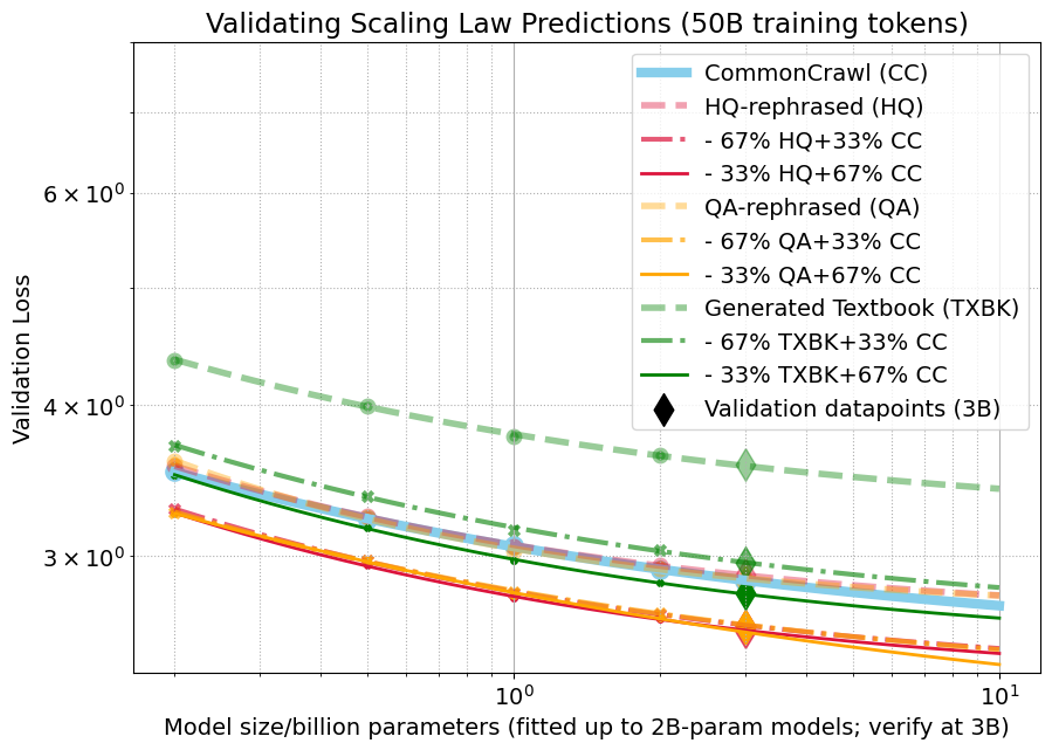}
        % \caption{Validation of the data scaling formula. Predictions for 200B tokens (fitted using up to 100B tokens) achieve an RMABE of 0.41\%. Solid dots display actual loss values while the fitted curves shows predicted loss. Validation datapoints are illustrated by diamond marks.}
        % \label{fig:dataval}
    \end{minipage}\hspace{-0em}
    \begin{minipage}{0.50\textwidth}
        \centering
        \includegraphics[width=\linewidth]{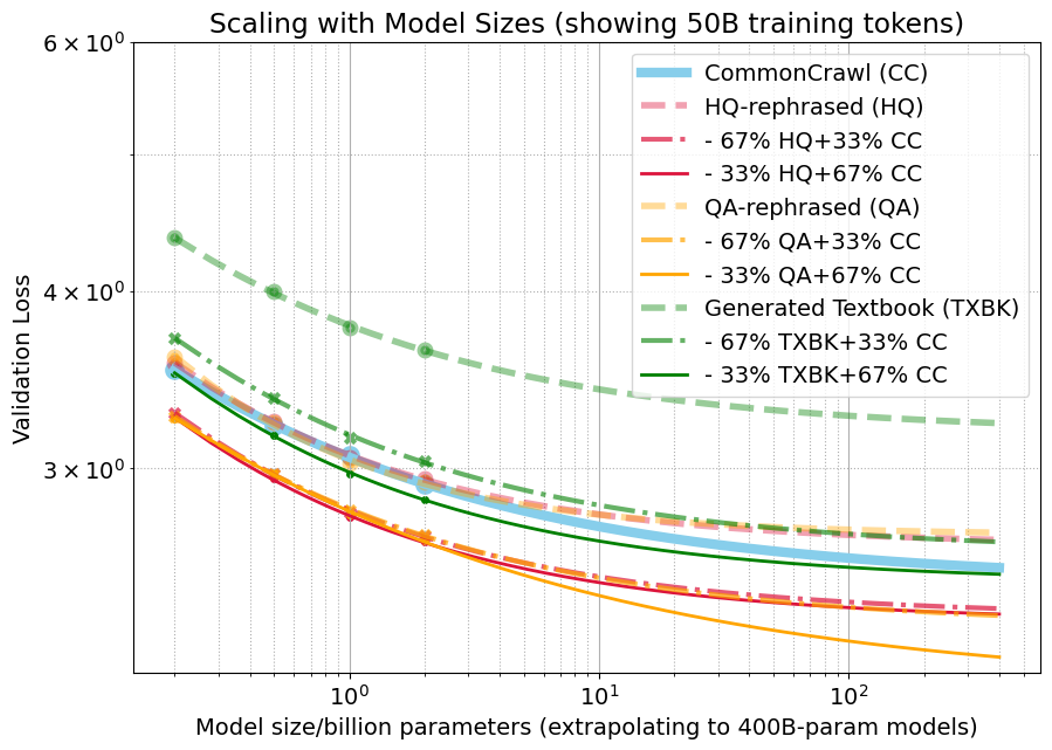}
        % \caption{Validation of the model scaling formula. Predictions for 3B-parameter models (fitted using up to 2B-parameter models) achieve an RMABE of 0.30\% on validation datapoints illustated with diamond marks.}
        % \label{fig:modelval}
    \end{minipage}
     \caption{Model Scaling. \textbf{Left:} Validation of the model scaling formula. Predictions for 3B-parameter models (fitted using up to 2B-parameter models) achieve an RMABE of 0.30\% on validation datapoints illustrated with diamond marks.  Solid dots display actual loss values while the fitted curves shows predicted loss. \textbf{Right:} Extrapolated model scaling performance for training on 50B tokens across various data mixtures.}
     \label{fig:model-valscale}
\end{figure}

\begin{enumerate}[leftmargin=*, itemsep=1pt, topsep=2pt]
        \item Pure synthetic data remains non-advantageous over CC; notably, models trained on \textbf{pure rephrased synthetic data will underperform those trained on CC at larger models.}
    \item For rephrased data mixtures, sensitivity to the mix ratio changes: while a 67\% synthetic mix was marginally better for larger \textit{data budgets} (data scaling), it becomes marginally \textit{disadvantageous} for larger \textit{model sizes} (model scaling) compared to a 33\% mix.
    \item For TXBK mixtures, 33\% synthetic consistently outperforms 67\%. The advantage of 33\% TXBK over pure CC \textbf{appears to diminish with larger models}, a trend not observed in data scaling.
\end{enumerate}

\textbf{Overall, these model scaling results suggest synthetic data appears \text{comparably} less favorable for pre-training larger LMs \text{relative to} its utility in data scaling scenarios.} Despite outperforming training on CC, larger models are not as tolerant to a higher ratio synthetic data as larger data budgets. This observation aligns with practices where synthetic data is effective for smaller LMs or specific pre-training phases, but less predominantly used for the largest models.

% \begin{figure}[htbp]
%     \centering
%     \includegraphics[width=0.7\textwidth]{figs/modelscale4.png}
%     \caption{Extrapolated model scaling performance for training on 50B tokens across various data mixtures.}
%     \label{fig:modelscale}
% \end{figure}

\subsubsection{Compute Scaling and Irreducible Loss}\label{ssec:jtscale}
We also fit joint scaling laws incorporating both model size ($N$) and data budget ($D$) using data from all $\sim$700 training runs (details in Appendix~\ref{sec:experiments}):
$
    \hat{\mathcal{L}}(N, D)=\frac{A}{N^\alpha}+\frac{B}{D^\beta}+E. 
$
An example loss landscape for CC data is shown in Fig.~\ref{fig:sljt}. The coefficient $E$ represents the \textit{irreducible loss}—the theoretical minimum loss achievable with infinitely large models and data.

\begin{figure}[htbp]
    \centering
    % Consider using a table for this data if the figure remains too busy, as suggested by \mk.
    % Alternatively, for the figure, ensure the legend is minimal or integrated, and color distinctions are clear.
    \includegraphics[width=0.7\textwidth]{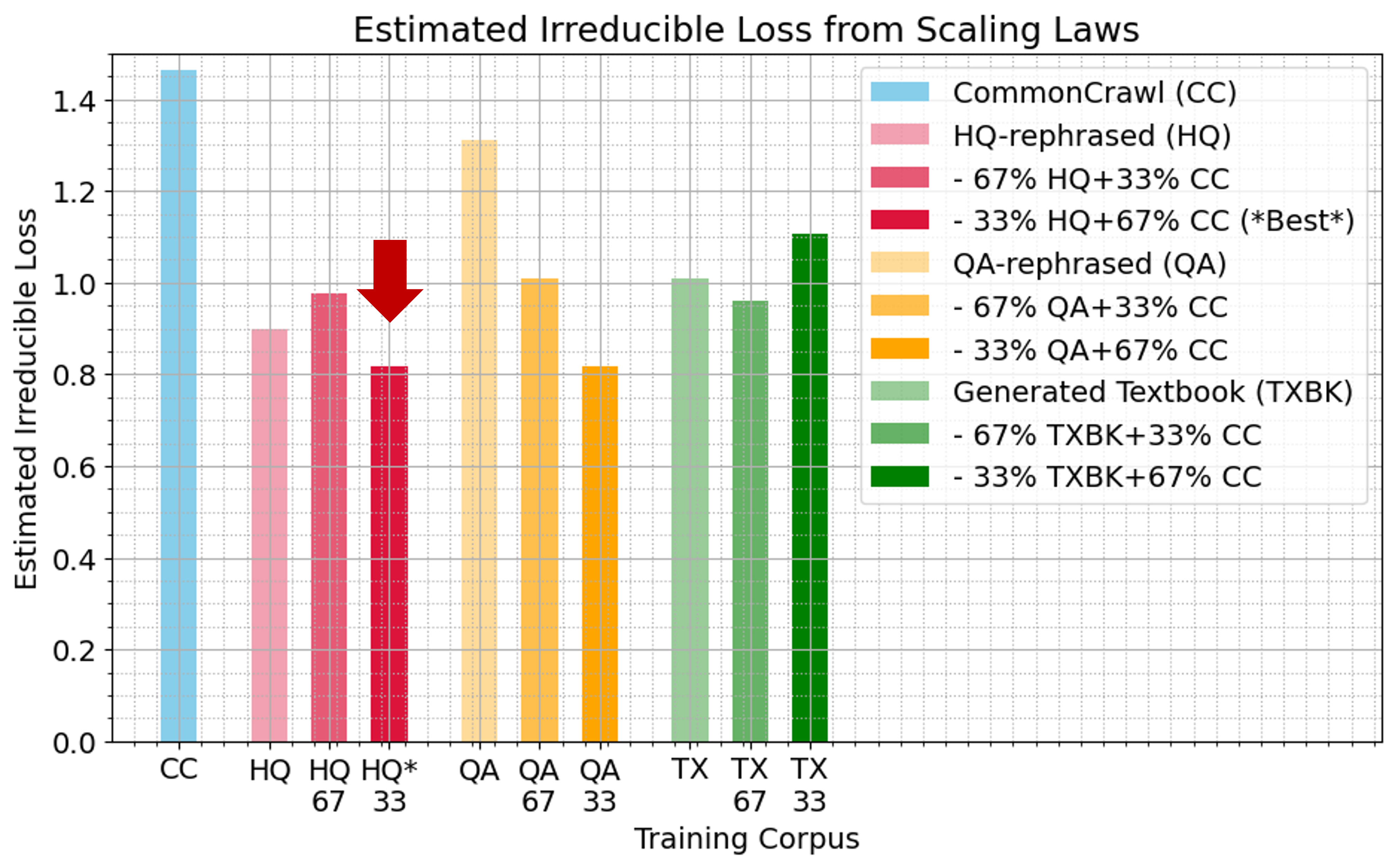}
    \caption{Estimated irreducible loss ($E$) for different data mixtures. Lower values are better.}
    \label{fig:irred}
\end{figure}

Estimations of $E$ for each data mixture (Fig.~\ref{fig:irred}) indicate their ultimate potential. Notably, \textbf{any mixture involving synthetic data, or pure synthetic data (except pure QA), is projected to achieve a lower irreducible loss than training only on CommonCrawl.} This empirically challenges theoretical concerns about ``model collapse'' \text{in single-round training,  which predict} any synthetic data inclusion would ultimately degrade performance \citep{dohmatob2024strong}.
Among the studied mixtures, \textbf{33\% HQ rephrased data + 67\% CC shows the lowest projected irreducible loss.} Conversely, pure QA rephrased data exhibits a high irreducible loss, second only to pure CommonCrawl.

\section{Additional Studies: A Broader View}
\label{sec:additional_studies}

Beyond the primary scaling law analysis, we conduct targeted experiments to deepen our understanding of specific factors influencing the effective use of synthetic data in pre-training.

\subsection{``Good'' Synthetic Data Mixture Ratios}
\label{ssec:optimal_ratio}

\paragraph{Motivation}
Our main scaling law analysis tested limited discrete mixture ratios (0\%, 33\%, 67\%, 100\%) of synthetic and natural data. To identify ``good'' ratios with finer granularity, we performed a fine-grained grid search, motivated by indications that optimal mixtures vary with synthetic data type, model scale, and data budget.

\paragraph{Methodology}
We trained approximately 400 additional LLMs (200M to 1B parameters) on data budgets from 1B to 50B tokens. For each synthetic data type (HQ, QA, TXBK) and each (model size, data budget) configuration, we varied the percentage of synthetic data mixed with CommonCrawl across ten exponentially spaced points: 0\%, 0.5\%, 1\%, 2\%, 5\%, 10\%, 15\%, 20\%, 50\%, and 100\%. The ``good'' ratio was defined as the mixture yielding the lowest validation loss on the evaluation sets.

\begin{figure}[h!]
    \centering
    \includegraphics[width=\textwidth]{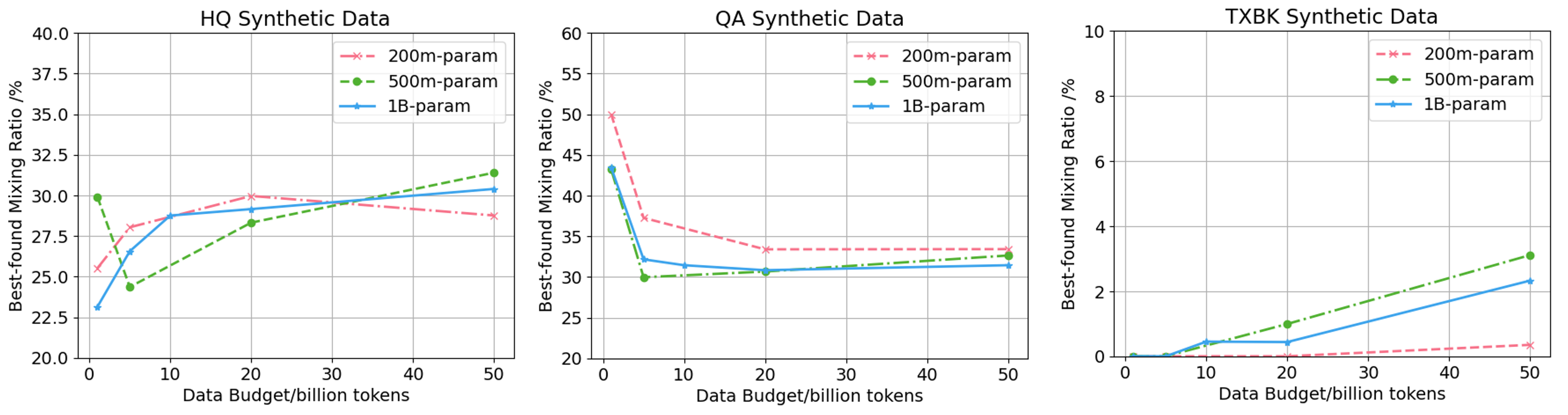} % 
    \caption{Best-found mixture ratios (percentage of synthetic data with CommonCrawl) from grid search for HQ (\textbf{Left}), QA (\textbf{Middle}), and TXBK (\textbf{Right}) synthetic data types across different model sizes and data budgets. Best-found ratios are all below $50\%$ appear to converge $\sim30\%$.}
    \label{fig:optimal_ratio_grid}
\end{figure}

\paragraph{Findings}
Figure~\ref{fig:optimal_ratio_grid} visualizes the results.  Best-found ratios are all below
50\% appear to converge $\sim$30\%. Key observations include:
\begin{itemize}[leftmargin=*, itemsep=1pt, topsep=2pt]
    \item \textbf{HQ Rephrased Data:} The optimal mixture is consistently \textbf{$\sim$30\% HQ synthetic data} with 70\% CommonCrawl across tested scales. This 30\% mixture generally outperformed the 50\% mixture suggested by \citet{maini2024rephrasing} in our setups.
    \item \textbf{QA Rephrased Data:} The preferred ratio of QA data tends to \textbf{decrease} with increasing model/data sizes, shifting from $\sim$50\% for smaller configurations towards 30\% for larger ones.
    \item \textbf{Textbook (TXBK) Data:} Benefits are most apparent at larger scales. Optimal ratios are often minimal (<5\%) for smaller configurations, increasing with scale but generally remaining below those for rephrased data.
\end{itemize}
These findings refine our scaling law observations, emphasizing the sensitivity of effective synthetic data deployment to its type and the training regime.

\subsection{Impact of Generator Model Capability}
\label{ssec:generator_ablation}

\paragraph{Motivation}
It is often assumed that larger, more capable generator models produce higher-quality synthetic data, leading to better downstream performance. We empirically tested this hypothesis.

\paragraph{Methodology}
We used Llama-3 models of varying scales (3B, 8B, and 70B parameters) as generators to recreate subsets of our HQ (High-Quality) and QA (Question-Answering) rephrased datasets. Generation prompts and source CommonCrawl documents were kept consistent with our original pipeline, which utilized Mistral-7B-Instruct as the generator.
A fixed 1B-parameter downstream model, with the same architecture as in previous experiments, was then trained for 5 billion tokens. For each generator (Llama3-3B/8B/70B), we evaluated the synthetic data produced by training the downstream model on mixtures with CommonCrawl. The percentage of synthetic data in these mixtures was varied across eight exponentially spaced points: 0.5\%, 1\%, 2\%, 5\%, 10\%, 15\%, and 20\%. Approximately 200 models were trained for this ablation study to compare the efficacy of synthetic data generated by models of different capabilities.

\begin{figure}[h!]
    \centering
    \includegraphics[width=\textwidth]{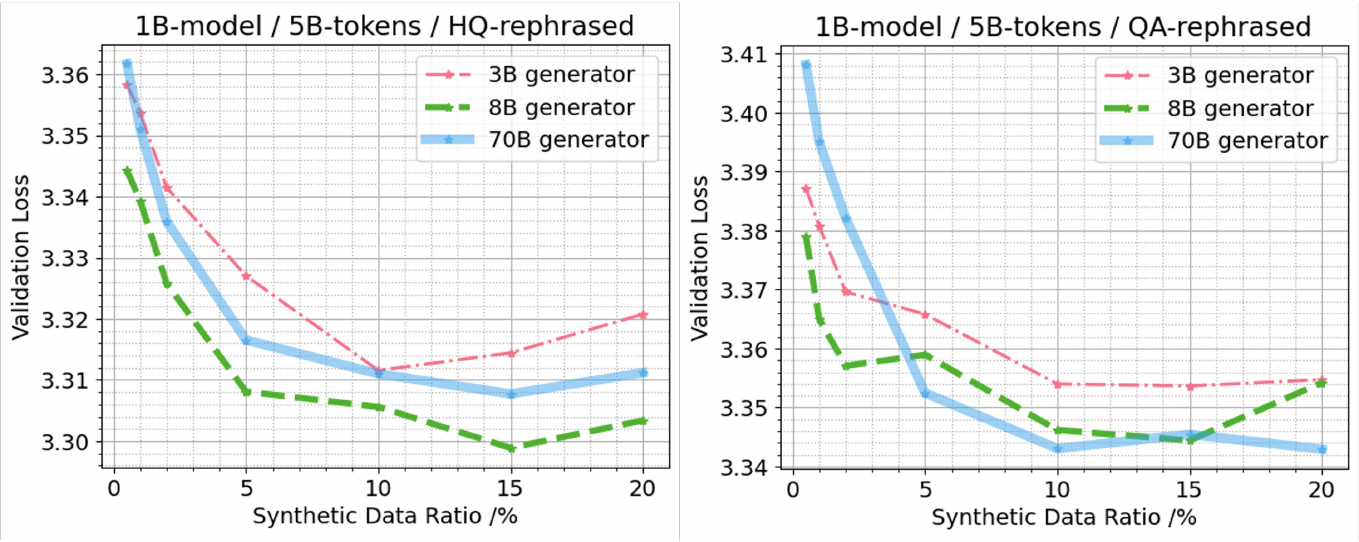} % TODO: Verify path
     \caption{\small{Generator model capability ablation. Compares validation loss of 1B-parameter models trained for trained for 5B tokens using mixtures of HQ/QA rephrased data from Llama3-3B/8B/70B generators with CommonCrawl. The percentage of synthetic data in these mixtures was varied across seven exponentially spaced points from 0.5\% to 20\%. }}\normalsize
    \label{fig:generator_ablation}
\end{figure}

\paragraph{Findings}
The results, illustrated by trends similar to those shown in Figure~\ref{fig:generator_ablation}, challenge the ``bigger is always better'' intuition for generator models and reveal a nuanced relationship:
\begin{itemize}[leftmargin=*, itemsep=1pt, topsep=2pt]
    \item A certain level of generator capability appears beneficial: synthetic data from Llama-3-8B generators consistently outperformed data from Llama-3-3B generators. \textbf{This finding suggests a baseline capability is necessary} and contrasts with suggestions from \citet{maini2024rephrasing}
    that rephrasing costs could be significantly reduced by using smaller generator LMs without a loss in downstream performance.
    \item However, \textbf{increasing generator size further to Llama-3-70B did not yield superior synthetic data} for pre-training compared to data from the Llama-3-8B generator, when assessed by the trained model's validation loss.
    \item In specific instances, the Llama-3-70B generator proved less effective. \textbf{For HQ rephrased data, synthetic data generated by Llama-3-70B models led to consistently worse} evaluation results than data from Llama-3-8B models. For QA rephrased data, the 70B generator's output resulted in comparable performance than that from the 8B generator.
\end{itemize}
This suggests that factors beyond sheer generator scale—such as instruction following fidelity at different scales, the diversity of generated outputs, or potential introduction of stylistic artifacts—play a crucial role in determining the utility of synthetic data for pre-training. Simply employing the largest available generator may not be the most effective or efficient strategy.

% \subsection{Interpreting the results via low-level statistics (n-gram analysis)} % Placeholder for future section

\subsection{Interpretation via Low-level Statistics}
\label{ssec:ngram}

The impact of synthetic data on pre-training efficiency, particularly how "good" mixing ratios vary with synthetic data type, budget, and model size, necessitates investigation into underlying mechanisms. While synthetic data may improve ``data quality'' (e.g., better coherence, reduced noise) at the cost of diversity \citep{long2024llms}, the generation process reflects the generator LM's output distribution, potentially shrinking the support of distribution from natural expressions or introducing artifacts like model collapse \citep{dohmatob2024strong}.

We investigate via low-level statistical analysis:
\begin{itemize} % [leftmargin=*, itemsep=1pt, topsep=2pt]
    \item[\textbf{(Q1)}]  Does synthetic data exhibit reduced lexical diversity (i.e., a "shrunk support") compared to natural web text?
    \item[\textbf{(Q2)}]  Can improvements in test performance with synthetic data be attributed to smaller training-test distributional distance?
    \item[\textbf{(Q3)}]  Are optimal mixing ratios due to minimized distributional distance or a more complex diversity-quality trade-off?
\end{itemize}
Inspired by \citet{magnusson2024paloma}'s finding that a small fraction of high-frequency strings contributes significantly to LM loss, we conduct unigram frequency analysis across training and test corpora.

Full results and analyses are provided in Appendix \ref{ssec:ngram}. We summarize the key findings:
\begin{itemize}[leftmargin=*]
    \item \textbf{Vocabulary Mismatches and High-Loss Tokens:} Unigrams frequent in test sets but rare or absent in some training sets (e.g., ``\verb|\n\n|'', `` hvor'' (Danish), `` dön'' (Turkish), etc.) cause higher evaluation loss. This issue is pervasive; no single training set offers complete coverage.
    \item \textbf{Synthetic Data and Unigram Distribution:} Synthetic data slightly shrinks unigram distribution compared to broad web corpora like CommonCrawl---yet CommonCrawl's wider coverage did not yield superior performance.
    \item \textbf{Distributional Distance to Test Sets:} KL-divergence on unigram distributions did not show synthetic data closer to test distributions; CommonCrawl appears closest to test datasets.
\end{itemize}

Preliminary conclusions are:
\begin{enumerate}[leftmargin=*, itemsep=1pt, topsep=2pt]
    \item \textbf{Inherent Limitations of Single Data Sources:} All data sources, including CommonCrawl, have distributional gaps causing high evaluation losses on underrepresented tokens. This favors mixed corpora with broad lexical coverage with reasonable frequencies, helping explain why mixed-source corpora often outperform single-source ones.

\item \textbf{Beyond Distributional Matching for ``Good'' Mixtures:} Models often train best with significant synthetic data proportions (e.g., $\approx$30\%) even if it does not minimize unigram distributional distance to test sets. This suggests factors beyond simple similarity, pointing to more complex diversity-quality trade-offs. % where the quality or structure benefits of synthetic data outweigh the slightly lower unigram diversity, enabling more efficient learning.
\end{enumerate}

\section{Conclusion}
\label{sec:conclusion}
This large-scale empirical investigation (over 1000 LLM variants) demonstrates that synthetic data in foundational pre-training presents a nuanced trade-off. Strategically mixing specific synthetic types (e.g., $\sim30\%$ high-quality rephrased text with natural data) can significantly accelerate pre-training convergence up to 5-10x and potentially achieve lower irreducible loss than natural data alone.
These results contribute mixed evidence on “model collapse" during large-scale single-round (n=1) model training on synthetic data--training on rephrased synthetic data shows no degradation in performance in foreseeable scales whereas training on mixtures of textbook-style pure-generated synthetic data shows patterns predicted by “model collapse".
However, effectiveness is also conditional on generation method, mixture strategy, and generator models. Larger generator models did not guarantee superior pre-training data. Downstream model performance cannot be simply explained by training data's diversity or similarity to test corpora, but pointing to more complex diversity-quality tradeoffs. 

Our findings underscore that synthetic data requires careful, empirically-informed deployment, rather than being a universal solution to data constraints. Essential \textbf{next steps} involve developing more targeted synthetic data generation techniques and dynamic mixing strategies. Rigorous evaluation of their long-term impacts on diverse model capabilities (reasoning, robustness, alignment) at frontier scales is crucial, alongside pinpointing key beneficial characteristics of generator models beyond sheer size.

\section{Discussions}
\paragraph{On "Model Collapse" for Large-Scale Single-Round (n=1) Model Training on Synthetic Data.}

This paper contributes new evidence on large-scale single-round (n=1) model training on synthetic data, rejecting certain conjectures from prior research on “model collapse" and helping refine their range of application.

The notion, “model collapse", was formalized by \citet{shumailov2023curse}, characterizing the effect of iterative training on self-generated (or mixed) data. Subsequent works such as \citet{dohmatob2024tale} studies the effect for n-fold iterative synthetic training, where the main results show training on synthetic data even for n=1 (one-round) leads to significant flare-up in test perplexity compared to training on the original data. Further, \citet{dohmatob2024strong} shows that even the smallest fraction of synthetic data (e.g., as little as 1\% of the total training dataset) can still lead to model collapse while training 124M-parameter GPT-2 small on BabyStories. Based on theoretical derivations, the authors conjecture that larger "models may mitigate the collapse, although they do not entirely prevent it." 
With strong conclusions on the important topic, the theoretical analysis is based on stylized models (e.g., regression models) and the language modeling experiments are simplistic (e.g., fine-tuning for one task). There remains a significant gap between these forecasts from “model collapse" and the advancement in generation/training on synthetic data.

This work brings more clarity to this evolving topic. In this work, we found that for one-round (n=1) model training on synthetic data:

\begin{itemize}[leftmargin=*, itemsep=1pt, topsep=2pt] 
    \item when using rephrased synthetic data in pre-training contemporary LMs, we do not see patterns of degradation in performance in foreseeable scales, and pre-training on rephrased synthetic data mixed with natural data can lead to significant speed-up in reducing validation loss.

\item training on mixtures of textbook-style pure-generated synthetic data did lead to notably higher loss on downstream domains, especially at small data budgets. This is largely consistent with the patterns and predictions reported in the “model collapse" papers.
\end{itemize}
This shows that in large-scale LM pre-training, training on synthetic data for one-round does not necessarily degrade validation performance, confining the extrapolation of theoretical results from “model collapse" papers.

\begin{itemize}[leftmargin=*, itemsep=1pt, topsep=2pt] 
    \item Despite the shrinking support on n-gram distributions, with the right type of synthetic data and a mixing ratio with natural data, the benefits for including synthetic data could outweigh the issues from “model collapse" and deliver substantial benefits. This adds counter evidence to the conjecture that including synthetic data would always lead to worse model performance.

\item However, most benefits observed are from rephrased synthetic data, whereas textbook-style synthetic data often leads to performance degradation even when mixed with a large proportion of natural data. Empirical results on textbook-style synthetic data show patterns characterized in "model collapse", suggesting the generalizability of theoretical results in “model collapse" may depend on the nature of synthetic data. 
\end{itemize}

Together, results and findings contributed in this work reject certain claims from “model collapse" and help refine their range of application.

\section*{Limitations}
\label{sec:limitations_concise}
Our study, while extensive, has limitations influencing the scope and generalizability of findings:
\begin{itemize}[leftmargin=*, itemsep=1pt, topsep=2pt] 
    \item \textbf{Limited Scope of Synthetic Data:} We analyzed three specific synthetic data types (HQ/QA rephrased, TXBK). Findings may not directly apply to other generation methods (e.g., synthetic code, dialogues) or prompting strategies.
    \item \textbf{Evaluation Focus:} Analysis heavily relied on perplexity/loss for scaling. It lacked in-depth human evaluations for nuanced capabilities or safety, and assessment on highly specialized tasks.
    Additional evaluations on NLP benchmarks would be a desirable addition as loss is not our final goal.
    
    \item \textbf{Temporal Effects:} We examined a single pre-training stage. Potential long-term effects, subtle degradation, or multi-generational dynamics (``model collapse'') were not investigated.
     \item \textbf{Scale Constraints:} Experiments reached 3B parameters and 200B tokens. Scaling trends observed regarding synthetic data utility at larger model sizes require validation at frontier model scales (>100B parameters, >10T tokens).
     \item \textbf{Impact of Tokenizers:} Studies in Section \ref{ssec:ngram} show that different training datasets have different coverage of tokens, where certain tokens rare in the training data may be associated with a higher loss in evaluation. Though not significant enough to affect the main results in this paper (such as the "good" mixing ratios), the impact of tokenizer may become more visible in finer-grained analysis on validation loss.
\end{itemize}

% \newpage

\clearpage
\section*{Ethical Considerations}
\label{sec:ethics_concise} % Optional label

The generation and use of synthetic data in LLM pre-training warrant careful ethical reflection:
\begin{itemize}[leftmargin=*, itemsep=1pt, topsep=2pt, itemsep=1pt, topsep=2pt] % Reduce spacing slightly
    \item \textbf{Bias Propagation:} Synthetic data risks inheriting and amplifying biases from generator models. Auditing generators and generated data for fairness is crucial but was outside this study's scope.
    \item \textbf{Factual Accuracy:} Generated content can include inaccuracies (hallucinations). Large-scale use could embed misinformation in models, necessitating robust quality control.
    \item \textbf{Data Diversity:} Over-reliance on potentially homogeneous synthetic data could reduce model robustness and diversity compared to training on varied real-world text.
    \item \textbf{Transparency \& Reproducibility:} We mitigate some concerns by committing to open-sourcing our full recipe to facilitate reproducibility and further community research.
\end{itemize}

\section*{Acknowledgment}
Ruoxi Jia and the ReDS lab acknowledge support through grants from the Amazon-Virginia Tech Initiative for Efficient and Robust Machine Learning, and the National Science Foundation under grants IIS-2312794, IIS-2313130, OAC-2239622.

\clearpage
\newpage
\bibliographystyle{assets/plainnat}
\bibliography{paper}

\clearpage
\newpage
\beginappendix

% \newpage
% \bibliography{custom}

\appendix

\section{Experimental Setup and Implementation Details}
\label{sec:experiments} % Kept label as in the provided draft, assuming it's referenced

% This content is from the draft's "\section{Experimental Setup and Implementations \todo{[under revision]}}"
% \subsection{Overview} % This subsection was commented out in the draft, so I'm omitting it.
% Our primary goal is to establish a controlled experimental framework to systematically compare the effects of natural versus synthetic data, and different synthetic data types, on LLM pre-training across various scales. We conduct large-scale pre-training experiments comparing models trained on: (1) a natural web corpus baseline, (2) purely synthetic datasets generated using distinct paradigms (Web Rephrasing and Synthetic Textbooks), and (3) various mixtures of natural and synthetic data. By training models of multiple sizes (ranging from 100M to 3B parameters) on these different data conditions across varying token budgets (ranging from 1B to 200B tokens) and evaluating performance comprehensively, we aim to analyze the effectiveness of each data type and rigorously study their scaling properties. We train approximately 600 LLM variants, consuming over 70,000 GPU hours on NVIDIA A100 80G hardware.

\subsection{Datasets}
\label{sec:datasets}

\subsubsection{Natural Data Baseline}
\label{sssec:natural_data}
Our natural data baseline consists of English text sourced from \textit{unfiltered CommonCrawl (CC) dumps processed via the RedPajama-v2 pipeline} \citep{weber2024redpajama}. This serves as our reference point representing widely used, large-scale web data.

\subsubsection{Synthetic Data Generation}
\label{sssec:synthetic_gen_impl_appendix} % Renamed label to avoid conflict if sssec:synthetic_gen_impl is in main text
To ensure consistency, a single generator model, \textit{Mistral-Instruct-7b-v0.1} \citep{jiang2023mistral}, was used for generating all synthetic datasets described below. Input documents for generation methods requiring a source were sampled from our unfiltered CommonCrawl baseline data. Standard sampling parameters (temperature=0.7, top-p=0.95) were generally used, unless otherwise specified (see Appendix~\ref{app:generation_specifics} for more details on prompts and post-filtering).

\paragraph{Method A: Web Rephrasing (WRAP-like)} Inspired by WRAP \citep{maini2024rephrasing}, we generated two stylistic variants by prompting Mistral-Instruct-7b-v0.1 to rewrite input CC documents (up to 2k tokens):
\begin{itemize}[leftmargin=*, itemsep=1pt, topsep=2pt]
    \item \textit{High-Quality (HQ) Rephrasing:} Used prompts optimized from the original WRAP work to produce longer ($\sim$550 tokens avg.) and more coherent synthetic texts, mimicking high-quality, well-structured English (e.g., Wikipedia style). The prompt aimed for clarity, coherence, and quality improvement while preserving core information.
    \item \textit{Question-Answering (QA) Rephrasing:} Used prompts optimized to restructure the input document's content into a conversational QA format ($\sim$550 tokens avg.), embedding instruction-following patterns.
\end{itemize}

\paragraph{Method B: Synthetic Textbooks (Phi-like)} Inspired by Phi \citep{li2023textbooks} and related community efforts (e.g., Cosmopedia \citep{benallal2024cosmopedia}), we generated novel textbook-style content. This involved a two-step process:
\begin{enumerate}[leftmargin=*, itemsep=1pt, topsep=2pt]
    \item An outline for a 10-chapter ``book'' was constructed based on keywords or themes extracted from input CC documents.
    \item Based solely on the outline, each chapter was generated ($\sim$450 tokens/chapter, $\sim$5k tokens/book), often including exercises and reference answers. We employed 4 prompt variations targeting different audiences (e.g., ``grade school students'', ``college students'', ``domain experts'', ``general audience'') to encourage diversity.
\end{enumerate}
Light heuristic filtering was applied post-generation to remove clearly malformed outputs (details in Appendix~\ref{app:generation_specifics}).

\subsubsection{Data Mixtures}
\label{sssec:data_mixtures_appendix} % Renamed label to avoid conflict
We created training datasets for various conditions by concatenating and shuffling the source data. For each synthetic data type (HQ, QA, Textbook), we prepared the following conditions relative to the unfiltered CC baseline:
\begin{itemize}[leftmargin=*, itemsep=1pt, topsep=2pt]
    \item 100\% Natural (Unfiltered CC)
    \item 100\% Synthetic (consisting entirely of one synthetic type: HQ, QA, or Textbook)
    \item 67\% Synthetic / 33\% Natural
    \item 33\% Synthetic / 67\% Natural
\end{itemize}
For each experimental point (model size, data budget), models were trained on these different mixture ratios corresponding to one synthetic type, allowing for direct comparison against the 100\% natural baseline. We typically trained 3-5 model variants per condition to enable robust scaling law analysis.

\subsection{Models and Training Configuration}
\label{ssec:models_training_appendix} 
\subsubsection{Model Architecture}
We employ a standard decoder-only Transformer architecture based on \textit{Llama 3 models} \citep{grattafiori2024llama}. Key architectural features include SwiGLU activation functions, RMSNorm for layer normalization, and Rotary Position Embeddings (RoPE). We train models at multiple sizes, \textit{ranging logarithmically from approximately 100 Million to 3 Billion parameters}, to facilitate scaling law analysis. 
In the research of scaling laws for large language models (LLMs),
model sizes are counted as \textit{non-embedding parameters}, which refers to all the \textit{learnable parameters }in the model except for those associated with the input and output token embeddings \citep{kaplan2020scaling}. This work follows this setup and the size of all models being trained refers to their count of non-embedding parameters. Table \ref{tab:model_architectures} outlines the architectural parameters for various model sizes explored.

\begin{table}[h!]
\centering
\caption{Model Architecture for Different Parameter Sizes}
\label{tab:model_architectures}
\begin{tabular}{lccc}
\toprule
Model  & Dimension & Layers  & Heads  \\
Size\quad\quad& ($d_{model}$)&($n_{layers}$)&($n_{heads}$)\\
\midrule
100M       & 576                   & 7                     & 9                   \\
200M       & 832                   & 10                    & 13                  \\
500M       & 1280                  & 16                    & 20                  \\
1B         & 1792                  & 22                    & 28                  \\
2B         & 2240                  & 28                    & 35                  \\
3B         & 2624                  & 32                    & 41                  \\
\bottomrule
\end{tabular}
\end{table}

\subsubsection{Training Hyperparameters}
\label{app:training_hyperparams} % Label for referencing from main text
All models were trained from scratch (random initialization) using the \textit{Meta Lingua library} \citep{meta_lingua} on PyTorch \citep{paszke2019pytorch} for efficient distributed training. We used the AdamW optimizer with $\beta_1=0.9, \beta_2=0.95$, a weight decay of 0.1. A cosine learning rate schedule was used with a linear warmup equivalent to 10\% of the total training steps. A consistent context length of 4096 tokens with an effective batch size of 1M tokens was used across all training runs. The Llama 3 pre-trained TikToken tokenizer with a 128k vocabulary size was used \citep{grattafiori2024llama}. Specific global batch sizes and gradient accumulation strategies were maintained consistently for comparable experimental settings to ensure fair comparisons. 
The models were trained using the hyperparameters detailed in Table \ref{tab:training_hyperparameters}.

\begin{table}[h!]
\centering
\caption{Training Hyperparameters}
\label{tab:training_hyperparameters}
\begin{tabular}{lr}
\toprule
Hyperparameter             & Value                         \\
\midrule
Optimizer                  & AdamW                         \\
Peak Learning Rate        & $3.0 \times 10^{-4}$          \\
Warmup Steps               & 10\%                        \\
Min. LR Ratio   & $1.0 \times 10^{-6}$          \\
Gradient Clipping    & 1.0                           \\
Batch Size (per device)    & 4                             \\
Accumulation Steps& 8                             \\
Sequence Length & 4096                          \\
(Effective Batch Size)     & \quad\,\,(1M tokens)                     \\
Prefetch Size              & 1024                          \\
Add BOS token              & True                          \\
Add EOS token              & True                          \\
Model Data Type            & bf16    \\
Epochs                     & 1                           \\
GPU Hardware               & \hspace{-1.5em}8x NVIDIA A100 80GB        \\
\bottomrule
\end{tabular}
\end{table}

Training time scales linearly with the training data size and scales nearly linearly with the model size (provided that it fits into GPU RAM without changing the batch size). Typically, we were able to train a 1B-parameter model for $\approx15$B tokens per day on a single node with 8x NVIDIA A100 80GB GPUs.

\subsubsection{Evaluation Protocol}
\label{sssec:evaluation_appendix} % Renamed label to avoid conflict

Model performance was evaluated using intrinsic metrics during training and a suite of downstream tasks post-training. Validation perplexity (log ppl equivalent to cross-entropy loss) was tracked during training. Final perplexity was calculated on a diverse set of \textit{14 non-code/math English text domains from The Pile} \citep{gao2020pile} (specifically: NIH ExPorter, Pile-CC, Wikipedia (en), USPTO Backgrounds, PubMed Central, PubMed Abstracts, PhilPapers, OpenWebText2, OpenSubtitles, Gutenberg (PG-19), FreeLaw, BookCorpus2, Books3, ArXiv) and also on the \textit{Wikitext-103} dataset \citep{merity2016pointer}. Per-token average perplexity across these domains serves as a key intrinsic performance measure. 
For model evaluation, the generator settings detailed in Table \ref{tab:evaluation_hyperparameters} were used.

\begin{table}[h!]
\centering
\caption{Hyperparameters for Perplexity Evaluation}
\label{tab:evaluation_hyperparameters}
\begin{tabular}{lr}
\toprule
Hyperparameter                 & Value \\
\midrule
Max Tokens to Generate       & 2048  \\
Generator Data Type (dtype)  & \quad\quad\quad\quad bf16  \\
\bottomrule
\end{tabular}
\end{table}

\section{Synthetic Data Generation: Prompts, Samples, and Parameters}
\label{app:generation_specifics}

This section provides further details on the synthetic data generation process, including the prompts used, sample outputs, and specific generation/filtering parameters.

\subsection{Prompt Templates for Synthetic Data Generation}
\label{app:prompts}
% TODO: Authors to insert specific prompts used for HQ Rephrasing, QA Rephrasing, and Synthetic Textbook chapter generation. 
% \todo{add our prompts and sample output from prev slides}
For HQ Rephrasing, prompts were adapted from \citet{maini2024rephrasing} and modified to encourage longer, high-quality text. For QA Rephrasing, prompts were designed to convert informational text into a question-answer dialogue format, adapted from \citet{maini2024rephrasing} and modified to promote better format-following and conversion for complete information. For Synthetic Textbooks, prompts guided the generation of chapter content based on outlines, with variations for different target audiences (grade school, college, expert, general).

% \begin{wrapfigure}{r}{0.5\textwidth}

\subsection{Sample Generations}
\label{app:samples}
% TODO: Authors to insert representative samples of generated HQ text, QA pairs, and Textbook excerpts.
This would ideally showcase the stylistic differences and typical output quality.

\subsection{Generation Parameters and Post-Filtering}
\label{app:generation_details} % Label for referencing from main text
All synthetic data was generated using a Mistral-Instruct-7b-v0.1 model.
\paragraph{Sampling Parameters:} Unless specified otherwise (e.g., for particular prompt explorations not detailed in the main paper), the following sampling parameters were used:
\begin{itemize}
    \item Temperature: 0.7
    \item Top-p (nucleus sampling): 0.95
\end{itemize}
These parameters were chosen to balance creativity and coherence in the generated text.

\paragraph{Post-Filtering:} Light heuristic post-filtering was applied to all generated synthetic datar, removing documents that were excessively short (e.g., less than 50 tokens) or excessively long relative to the target length for that generation type, if such outputs occurred despite prompt length guidance. The goal of this light filtering was to remove egregious generation errors without overly sanitizing the data or significantly altering its distribution.

% non-embedding parameters (exclude the oputput 
% In the context of scaling laws for large language models (LLMs), non-embedding parameters refer to all the learnable parameters in the model except for those associated with the input token embeddings and, sometimes, the output token (un-)embeddings. 
% \citep{kaplan2020scaling}

% model:
%     dim: 1792
%     n_layers: 22
%     n_heads: 28
% modeldic = {'3b':['2624','32','41'], '2b':['2240','28','35'], '1b':['1792','22','28'], '500m':['1280','16','20'], '200m':['832','10','13'], '100m':['576','7','9']}",
%     "

%     training time

%     token rate

%     optim:
%     lr: 3.0e-4
%     warmup: 2000
%     lr_min_ratio: 0.000001
%     clip: 1.0

%     batch_size: 4
%     prefetch_size: 1024 
%     seq_len: 4096

%         add_bos: true
%     add_eos: true

%         model_dtype: bf16

%         eval:
%     generator:
%         max_tokens: 2048
%         dtype: bf16

% TODO: Authors to add further specific architectural details of the Llama 3-based models if necessary,
% beyond what is mentioned (SwiGLU, RMSNorm, RoPE). This could include number of layers/heads for each model size,
% hidden dimensions, intermediate FFN sizes, etc.

\section{Supplementary Discussion on Related Work}
\label{app:extended_related_work_appendix} % Changed label from sec:appendix
% This content is from the draft's "\section{Extended Related Work}"

This appendix provides supplementary details to the related work discussed in Section~\ref{sec:related_work}, offering further context on synthetic data applications, data curation practices, and model collapse theories.

\subsection{Synthetic Data in Post-training} % Using letters for top-level points in this specific appendix section for differentiation
The use of synthetic data is particularly well-established and successful in post-training phases, primarily for aligning LLMs with human instructions and preferences. This success stems from the ability to generate large amounts of targeted data for specific, often narrow, objectives where human annotation is costly or slow. Key examples include:

\begin{minipage}{\textwidth}
\begin{tcolorbox}[title=\small{B.1.1 Prompt Template HQ Rephrasing},colback=lightgray!30!white,colframe=black]
\small{\begin{itemize}[leftmargin=*]
    \item \textbf{System Prompt:} Provide direct and detailed response to the instructions without adding additional notes.
    \item \textbf{[USER]:} For the following document, regardless of its original content or formatting, write a full article of the same content in high quality English language as in texts on Wikipedia: [xxxx]. Provide the rephrased article without any additional notes. Long article with full length and complete details. Rephrased article:
\end{itemize}}
\end{tcolorbox}
% \end{wrapfigure}
\begin{tcolorbox}[title=\small{B.1.2 Prompt Template QA Rephrasing},colback=lightgray!30!white,colframe=black]
\small{\begin{itemize}[leftmargin=*]
    \item \textbf{System Prompt:} Provide direct and detailed response to the instructions without adding additional notes.
    \item \textbf{[USER]:} For the following document, regardless of its original content or formatting, convert it into a comprehensive list of question-answer pairs with multiple tags of ``Question:'' followed by ``Answer:'', where questions and answers cover complete information of the original document. Document: [xxxx].  Provide the converted question-answer pairs without any additional notes. Question-answer pairs with corresponding tags (``Question:'', ``Answer:''):
\end{itemize}}
\end{tcolorbox}

\begin{tcolorbox}[title=\small{B.1.3 Prompt Template for Generating Textbook-style Synthetic Data: Step 1, Outline Generation},colback=lightgray!30!white,colframe=black]
\small{\begin{itemize}[leftmargin=*]
    \item \textbf{Step 1: generate an outline based on input text.} 
    \item \textbf{System Prompt:} Provide direct and detailed response to the instructions without adding additional notes. 
    \item \textbf{[USER] <4 versions>: } Imagine you are a prolific author tasked with writing a textbook. You are working on writing a textbook involving the knowledge and information of the following text. Text: [xxxx]\textbackslash n Your task is to write an outline for the textbook. Your target audiences are <grade school students/college students/field experts/general public>. The textbook has 10 chapters in total plus title, introduction, and appendices. Textbook outline:
\end{itemize}}
\end{tcolorbox}
\begin{tcolorbox}[title=\small{B.1.4 Prompt Template for Generating Textbook-style Synthetic Data: Step 2, Chapter Generation},colback=lightgray!30!white,colframe=black]
\small{\begin{itemize}[leftmargin=*]
\item \textbf{Step 2: generate each section based on outline.} 
    \item \textbf{System Prompt:} Provide a direct and detailed response to the instructions without adding additional notes. 
    \item \textbf{[USER]: } Imagine you are a prolific author tasked with writing a textbook. You are working on writing a textbook with the following outline.\textbackslash n Outline: [xxxx] \textbackslash n Your task is to write Chapter x of the textbook. Your target audiences are grade school students. Include exercises at the end of the chapter to test the reader's knowledge of the chapter and then provide reference answers to each question.
\end{itemize}}
\end{tcolorbox}    
\end{minipage}

\begin{tcolorbox}[title=\small{B.2.1 Sample Generation for HQ Rephrasing},colback=lightgray!30!white,colframe=black]
\small{\begin{itemize}[leftmargin=*]
    \item \textbf{[Input text]:} Porsche® 911 B\&B Hi-Capacity Intercooler, 1978-1989 Home
    
/ Porsche® 911 B\&B Hi-Capacity Intercooler, 1978-1989

Porsche® 911 B\&B Hi-Capacity Intercooler, 1978-1989 More Views

The B \& B intercooler offers 100\% more cooling area than the stock unit, and 35\% more than any other aftermarket cooler. Due to this increased size, the cooler has greater thermal efficiency, which means cooler air to the engine. This will reduce the chance of detonation and allow for greater boost pressure for more power. Bolts in stock location with minor modifications. Choose from Silver Aluminum or Black Powder Coated (allow 2-weeks). APPLICATIONS

PORSCHE 911 (930) 1978, 1979, 1980, 1981, 1982, 1983, 1984, 1985, 1986, 1987, 1988, 1989

    \item \textbf{HQ Rephrased:} The Porsche® 911 B\&B Hi-Capacity Intercooler is a high-performance cooling system designed for the Porsche 911 (930) model from 1978 to 1989. This intercooler offers 100\% more cooling area than the stock unit, and 35\% more than any other aftermarket cooler. This increased size allows for greater thermal efficiency, which means cooler air to the engine. This, in turn, reduces the chance of detonation and allows for greater boost pressure, resulting in more power.
    
The B\&B intercooler is designed to bolt into the stock location with minimal modifications. It is available in two finishes: Silver Aluminum and Black Powder Coated. Customers should allow 2-weeks for delivery of the Black Powder Coated finish.

The applications for the Porsche® 911 B\&B Hi-Capacity Intercooler are the Porsche 911 (930) models from 1978 to 1989. This intercooler is designed to provide superior cooling performance, making it an ideal choice for those looking to improve the performance of their Porsche 911.

\end{itemize}}
\end{tcolorbox}

\begin{tcolorbox}[title=\small{B.2.2 Sample Generation for QA Rephrasing},colback=lightgray!30!white,colframe=black]
\small{\begin{itemize}[leftmargin=*]
    \item \textbf{[Input text]:} Porsche® 911 B\&B Hi-Capacity Intercooler, 1978-1989 Home
    
/ Porsche® 911 B\&B Hi-Capacity Intercooler, 1978-1989

Porsche® 911 B\&B Hi-Capacity Intercooler, 1978-1989 More Views

The B \& B intercooler offers 100\% more cooling area than the stock unit, and 35\% more than any other aftermarket cooler. Due to this increased size, the cooler has greater thermal efficiency, which means cooler air to the engine. This will reduce the chance of detonation and allow for greater boost pressure for more power. Bolts in stock location with minor modifications. Choose from Silver Aluminum or Black Powder Coated (allow 2-weeks). APPLICATIONS

PORSCHE 911 (930) 1978, 1979, 1980, 1981, 1982, 1983, 1984, 1985, 1986, 1987, 1988, 1989

    \item \textbf{QA Rephrased:} Question: What is the name of the product?
    
Answer: Porsche® 911 B\&B Hi-Capacity Intercooler, 1978-1989

Question: What is the cooling area of the stock unit?

Answer: 100\% more cooling area than the stock unit

Question: What is the cooling area of the B\&B intercooler compared to other aftermarket coolers?

Answer: 35\% more than any other aftermarket cooler

Question: What is the purpose of the increased size of the cooler?

Answer: Greater thermal efficiency, which means cooler air to the engine

Question: What effect will this have on the engine?

Answer: Reduced chance of detonation and allow for greater boost pressure for more power

Question: Where will the bolts for the cooler be installed?

Answer: Bolts in stock location with minor modifications

Question: What are the two options for the color of the cooler?

Answer: Silver Aluminum or Black Powder Coated

Question: How long will it take for the cooler to arrive?

Answer: 2-weeks for Black Powder Coated

Question: What are the applications of the B\&B intercooler?

Answer: Porsche 911 (930) 1978, 1979, 1980, 1981, 1982, 1983, 1984, 1985, 1986, 1987, 1988, 1989.

\end{itemize}}
\end{tcolorbox}

\begin{tcolorbox}[title=\small{B.2.3 Sample Generation for Textbook-style Synthetic Data: Step 1, Outline Generation},colback=lightgray!30!white,colframe=black]
\small{\begin{itemize}[leftmargin=*]
    \item \textbf{[Sample outline <general public>]:}  Title: The Baltimore Orioles' Spring Training
    
Introduction:

* Brief overview of the Orioles' history and recent successes

* Importance of spring training in preparing for the regular season

* Purpose of the textbook: To provide readers with an in-depth look at the Orioles' spring training activities

Chapter 1: Overview of the Orioles' Spring Training Facilities

* Description of the team's two main spring training fields: Ed Smith Stadium and the Gulf Coast League

* Explanation of the new artificial turf field at Ed Smith Stadium

* Importance of having multiple fields for inclement weather and rehabbing injured players

Chapter 2: Injury Updates and Rehabilitation

* Overview of the Orioles' key injuries and their status during spring training

...

Chapter 10: The Future of Orioles Baseball

* Overview of the team's current roster and their potential for success in the upcoming season

* Description of the team's prospects and their potential to develop into top Major League players

* Explanation of the importance of continued investment in player development and team building for long-term success

Conclusion:

* Recap of the key takeaways from the Orioles' spring training

* Final thoughts on the importance of spring training and the team's potential for success in the upcoming season

Appendices:

* List of Orioles' major league roster and key prospects

* Glossary of baseball terms and concepts discussed in the textbook

* Additional resources for further reading on baseball and the Orioles.
\end{itemize}}
\end{tcolorbox}
\paragraph{Instruction Generation (Self-Instruct):} Techniques like Self-Instruct \citep{wang2022self} use an LLM to bootstrap instruction-following data (instruction, input, output tuples) from a small seed set, enabling effective instruction fine-tuning without extensive human labeling.

\paragraph{Preference Data for RLHF:} Reinforcement Learning from Human Feedback (RLHF) requires data indicating preferences between model outputs. Synthetic data generation can augment or replace human effort in creating these preference pairs, facilitating large-scale alignment \citep{alpaca}.

\paragraph{AI Feedback and Constitutional AI:} Methods like Reinforcement Learning from AI Feedback (RLAIF) and Constitutional AI \citep{bai2022constitutional} utilize AI models themselves, guided by principles or rules, to provide feedback or generate preferred responses, further automating the alignment process.

\subsection{Data Curation Practices}
The quality and composition of pre-training data are paramount. Major dataset development efforts highlight common practices:

\paragraph{Large-Scale Corpora Examples:}
\begin{itemize}[leftmargin=*]
    \item C4 (Colossal Clean Crawled Corpus) \citep{raffel2020exploring}: Derived from Common Crawl using heuristics like language filtering (retaining primarily English text), removing code/markup, filtering based on blocklists, and deduplication at the document level.
    \item The Pile \citep{gao2020pile}: A diverse dataset combining 22 different high-quality sources, including academic papers (PubMed, arXiv), books (Books3), code (GitHub), web text (Pile-CC), and conversational data, with source-specific filtering.
    \item RefinedWeb \citep{penedo2024fineweb}: Focused on rigorous filtering and aggressive fuzzy deduplication of web data from Common Crawl to create a high-quality, large-scale web corpus, arguing against heuristic domain mixing.
\end{itemize}

\paragraph{Domain Mixing:} Research actively explores the impact of mixing data from different sources \citep{liu2024regmix,xie2023doremi,kang2023performance}. For example, including code data \citep{touvron2023llama} or synthetic reasoning data \citep{abdin2024phi} has been shown to improve reasoning, while the optimal ratio of web text, books, and other domains may vary depending on evaluation metrics and model scale \citep{ye2024data,kang2024autoscale,kang2024get}.

\subsection{Model Collapse Mechanisms}
The theoretical concern of model collapse \citep{shumailov2023curse, dohmatob2024tale, dohmatob2024strong} posits that training generative models on their own output can lead to degenerative feedback loops. Proposed mechanisms include:
% Citations [34-40] from the original draft text were broad; replaced with more specific ones if known or kept general.

\textit{Distributional Drift:} The distribution of synthetically generated data may subtly differ from the true underlying data distribution. Iterative training can amplify these differences, causing the model's learned distribution to drift further away.

\textit{Loss of Diversity:} Models might over-represent common modes in the data they generate, leading to a gradual loss of information about less frequent phenomena or the tails of the distribution (``tail-forgetting'').

\textit{Artifact Amplification:} Flaws, biases, or stylistic quirks of the generator model may be replicated and amplified in subsequent generations trained on its output.
Understanding the empirical conditions under which these theoretical risks manifest in large-scale LLM training is an ongoing research effort.

\section{Additional Experiment Results}
We provide some additional results and analyses to 
experiments and studies in Sections \ref{sec:empirical_results} and \ref{sec:additional_studies}.
\subsection{Additional Results on Section \ref{ssec:jtscale}}
\begin{figure}[htbp]\vspace{-0em}
    \centering
    \includegraphics[width=0.48\textwidth]{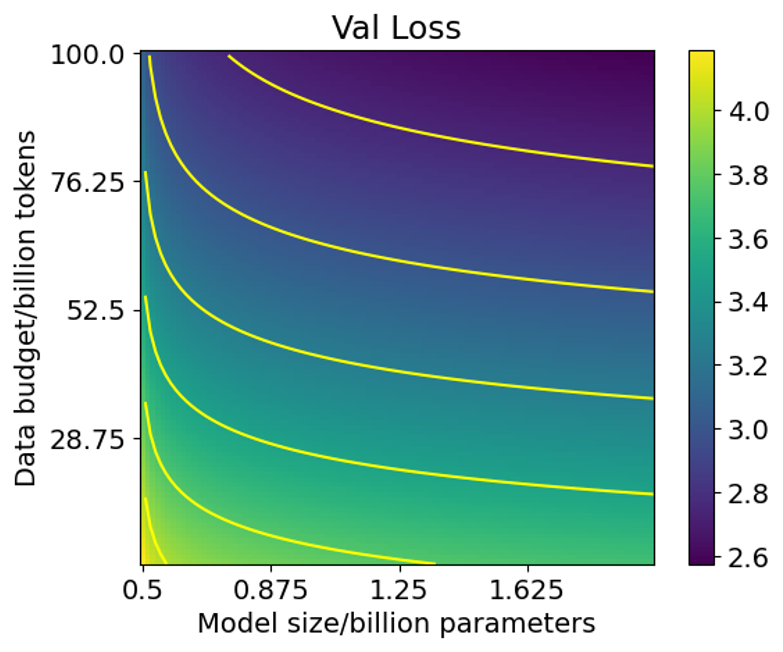}
    \caption{Joint fitted scaling law predictions for models trained only on CommonCrawl (CC). Yellow lines are loss contours.}
    \label{fig:sljt}\vspace{-1em}
\end{figure}

Figure \ref{fig:sljt} shows joint fitted scaling law predictions for models trained only on CommonCrawl (CC) with visualization for loss contours.

% \subsection{Additional Results on Section \ref{ssec:generator_ablation}}

% Figure \ref{fig:optimal_ratio_grid} shows best-found mixture ratios (percentage of synthetic data with CommonCrawl) from grid search for HQ, QA, and TXBK data types across different model sizes and data budgets. 
% \begin{figure}[h!]
%     \centering
%     \includegraphics[width=0.45\textwidth]{figs/allratio2.png} % TODO: Verify path
%     \caption{Best-found mixture ratios (percentage of synthetic data with CommonCrawl) from grid search for HQ, QA, and TXBK data types across different model sizes and data budgets. Best-found ratios are all below $50\%$ appear to converge $\sim30\%$.}
%     \label{fig:optimal_ratio_grid}\vspace{-1em}
% \end{figure}

\newpage
\clearpage
\subsection{Additional Results on Section \ref{ssec:ngram}}
\begin{figure}[h]
    \centering
    % Two subfigures side by side (top)
    \begin{subfigure}[b]{0.5\textwidth}
        \centering
        \includegraphics[width=\textwidth]{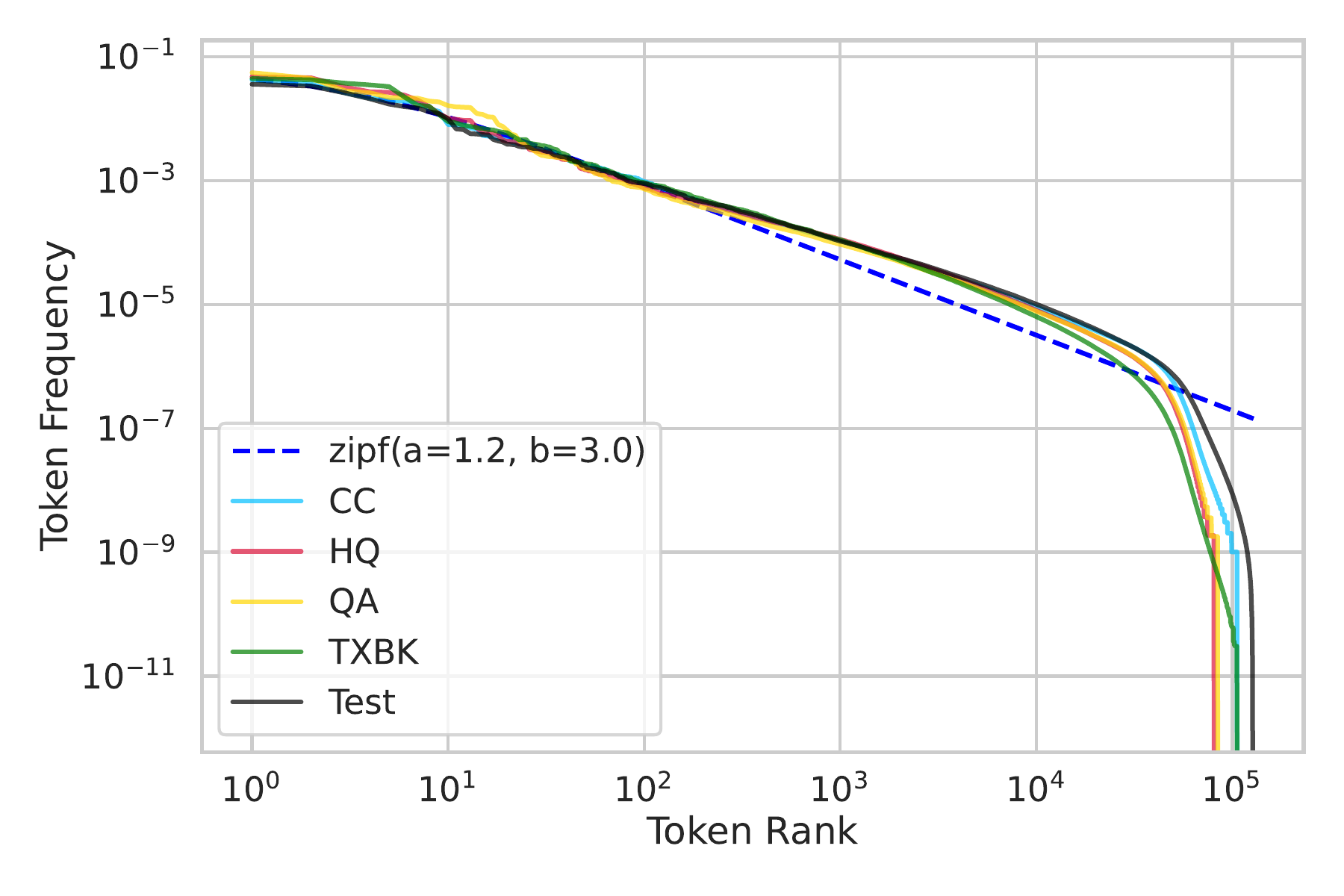}
        \caption{Unigram analysis with Zipf function fitting for unigram (token) frequencies of different training data corpora. CC appears to have wider and slight more uniform coverage of tokens than other training corpora, whereas the test corpora have wider coverage of tokens than training corpora.}\label{fig:zipf_cc}
    \end{subfigure}%
    \hspace{1em}
    \begin{subfigure}[b]{0.45\textwidth}
        \centering
        \includegraphics[width=\textwidth]{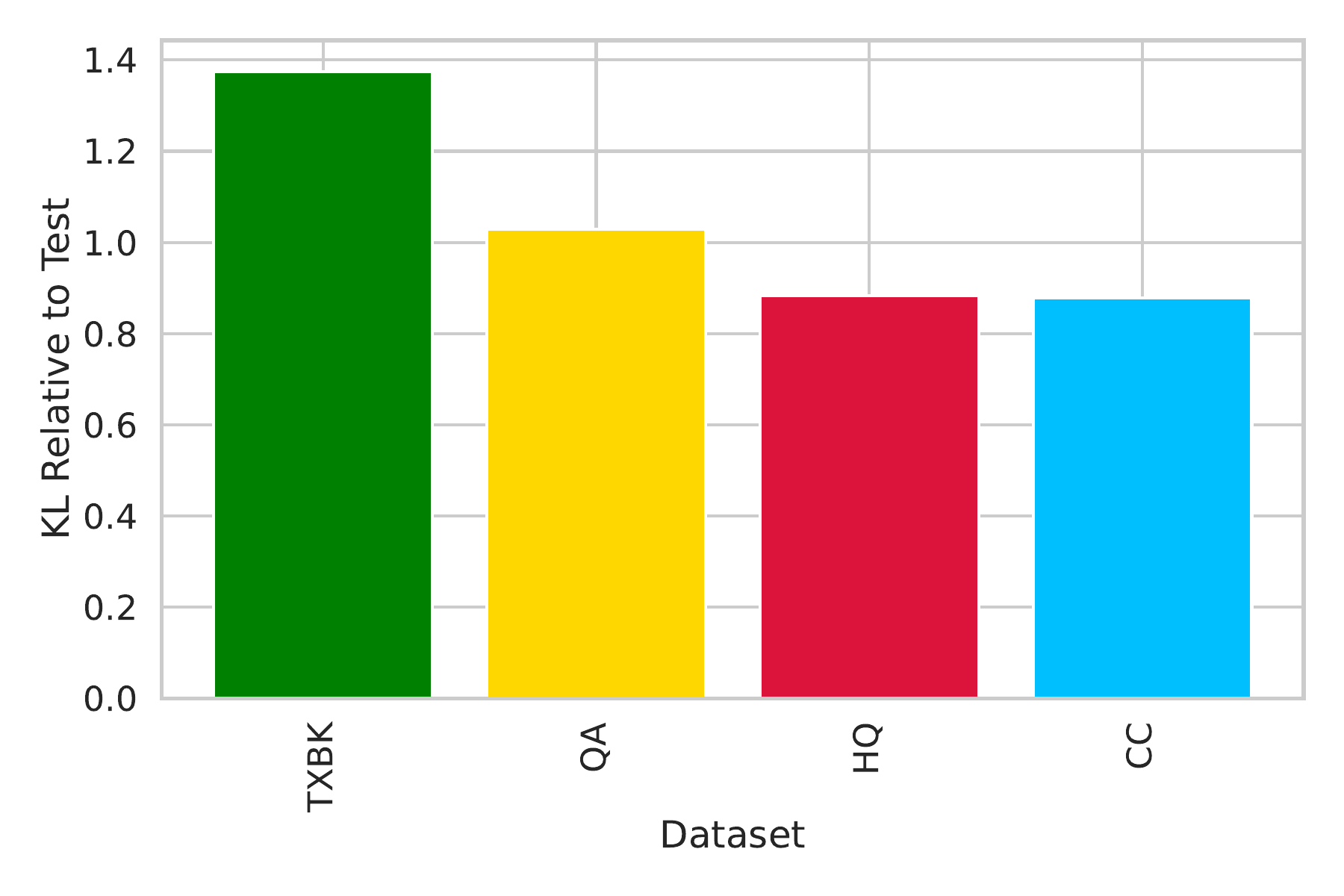}
    \caption{KL-divergence beween unigram distributions of the test dataset and each training corpus. CC appears to have the smallest KL-divergence from test data, suggesting the highest distribution similarities, but does not yield high dowmstream model performance.}
    \label{fig:kl}
    \end{subfigure}
    \vspace{1em} % Adjust vertical spacing accordingly
    
    % Larger, bottom figure spanning entire width
    \begin{subfigure}[b]{\textwidth}
        \centering
        \includegraphics[width=\textwidth]{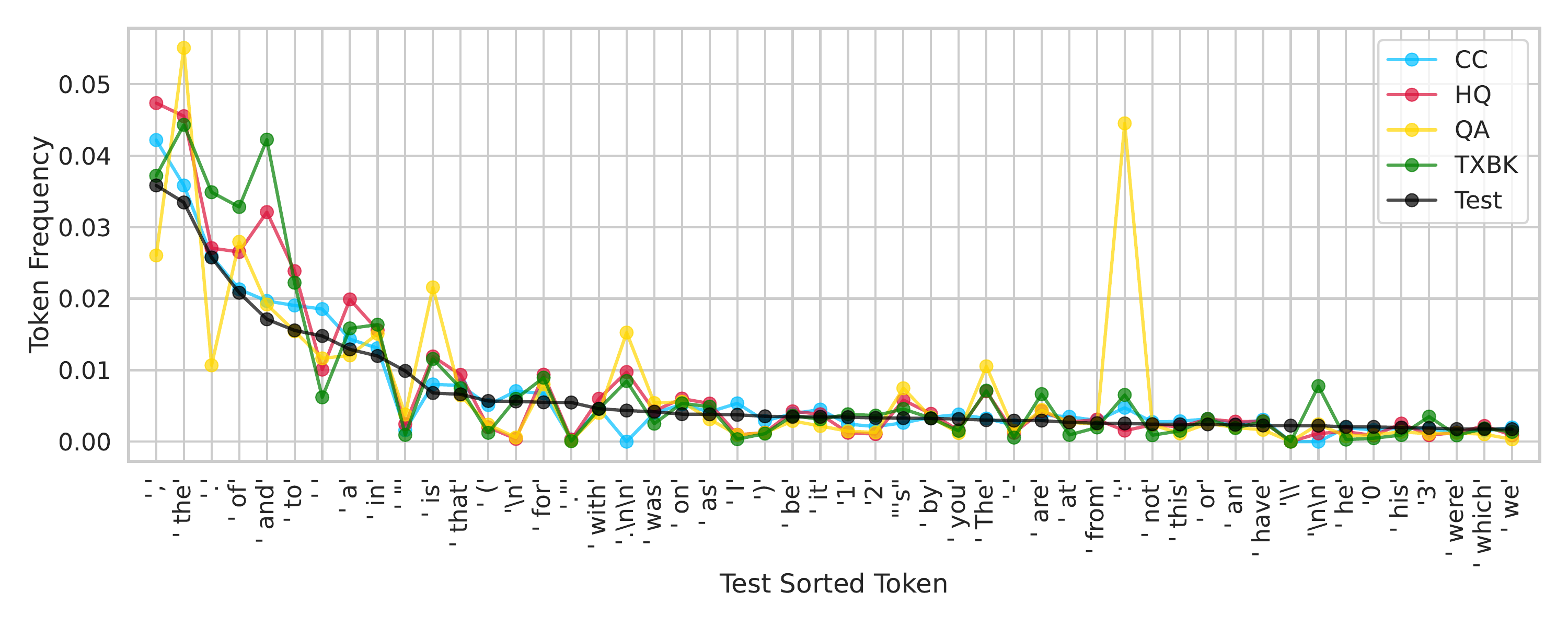}
        \caption{All estimated token frequencies for each training and test corpus, sorted by the frequency of test tokens. Some methods have relatively lower representation of certain tokens (e.g., CC's ``\texttt{.\char`\\n\char`\\n}'') and others increase certain token frequencies (e.g., QA's ``\texttt{:}'').}\label{fig:all_prob}
    \end{subfigure}

\caption{Analysis of token distributions across datasets and methods.}
\label{fig:overall fig}
\end{figure}

\begin{sidewaysfigure}
    \centering
    \includegraphics[width=\textwidth,height=\textheight,keepaspectratio]{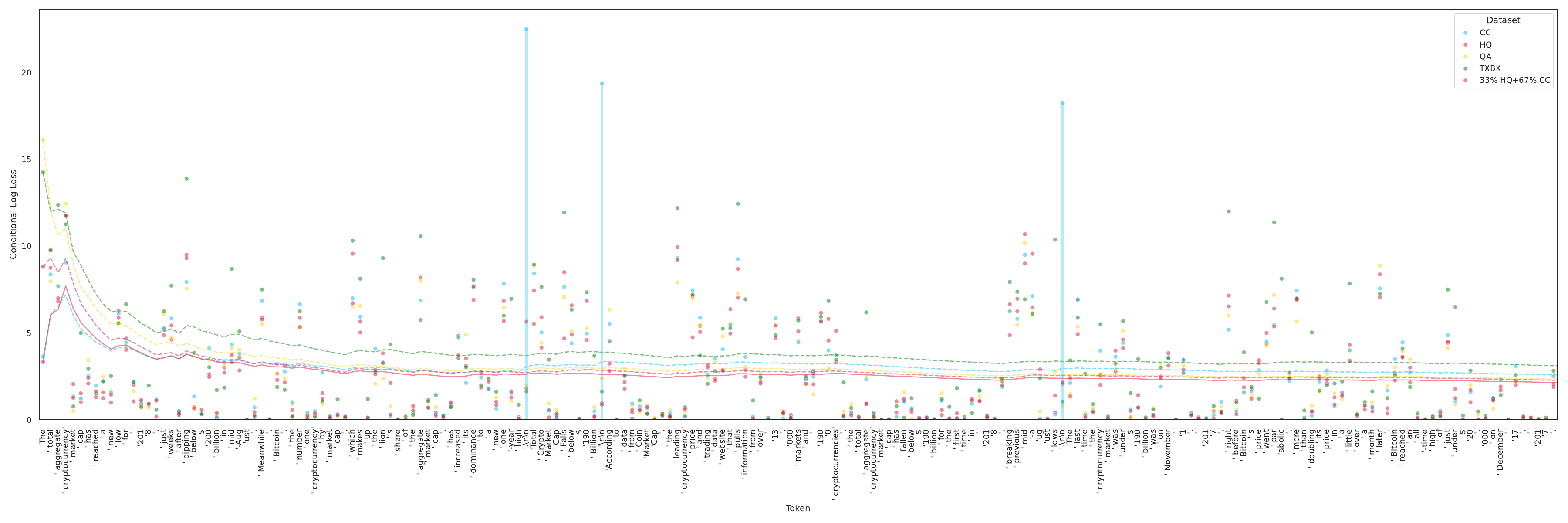}
    \caption{Visualization of per-token loss (dots) and rolling average (curves) in evaluation for 1B-parameter models trained for 100B tokens on respective corpus. The model trained on CC, which has a low frequency for the token ``\texttt{.\char`\\n\char`\\n}'', shows high loss in evaluation when encountering this token, though the rolling average remains stable.}
    \label{fig:rotated_fullpage}
\end{sidewaysfigure}

Figure \ref{fig:overall fig} shows analyses of token distributions across datasets and methods. Figure \ref{fig:rotated_fullpage} visualizes per-token loss and rolling avergage in evaluation for models trained on respective corpus. 

\label{app:ngram}
% Overall figure environment
    
% \begin{wrapfigure}{r}{1.0\textwidth}
% \subsection{additional results on low level statistics}
%     \centering
%     \vspace{1em}

% \begin{minipage}{0.4\textwidth}
%         \centering
%         \includegraphics[0.4\textwidth]{figs/cleaned_zipf_fit_cc.pdf}
%             \caption{zip-f}
%             \label{fig:zipf}
%     \end{minipage}
%     \begin{minipage}{0.4\textwidth}
%         \centering
%         \includegraphics[0.4\textwidth]{figs/cleaned_zipf_fit_cc.pdf}
%             \caption{zip-f}
%             \label{fig:zipf}
%     \end{minipage}
% \vspace{1em}

%     \includegraphics[width=\textwidth]{figs/all_probabilities_clean_test_summary_50.pdf}
%     \caption{All estimated token frequencies, sorted by top test tokens. Some methods have relatively lower representation of certain tokens (e.g., CC's ``\texttt{.\char`\\n\char`\\n}'') and others increase certain token frequencies (e.g., QA's ``\texttt{:}'').}
%     \label{fig:all prob}\vspace{-5em}
% \end{wrapfigure}

% \begin{figure*}[h!]
%     \centering\vspace{1em}
%     \includegraphics[width=\textwidth]{figs/all_probabilities_clean_test_summary_50.pdf}
%     \caption{All estimated token frequencies, sorted by top test tokens. Some methods have relatively lower representation of certain tokens (e.g., CC's ``\texttt{.\char`\\n\char`\\n}'') and others increase certain token frequencies (e.g., QA's ``\texttt{:}'').}
%     \label{fig:all prob}\vspace{-5em}
% \end{figure*}
\clearpage

\end{document}